\documentclass{ifacconf_amx}
\usepackage{times}
\usepackage{latexsym}
\usepackage[utf8]{inputenc}
\usepackage{natbib}
\usepackage{booktabs} 
\usepackage{microtype}
\usepackage{graphicx}
\usepackage{placeins} 
\usepackage{float}    

\usepackage{xcolor}
\usepackage[most]{tcolorbox}

\definecolor{myblue}{RGB}{0,0,255} 

\newcommand{\highlightblue}[1]{\textcolor{myblue}{\texttt{\detokenize{#1}}}}

\tcbuselibrary{breakable} 
\newtcolorbox{promptbox}[1][]{
  breakable,
  colback=white, 
  colframe=black, 
  arc=3mm, 
  boxrule=0.5pt, 
  fonttitle=\bfseries,
  fontupper=\small, 
  title=#1, 
  left=4mm,
  right=4mm,
  top=2mm,
  bottom=2mm
}

\newtcolorbox{appendixpromptstyle}{
  width=\linewidth,
  colback=white,
  colframe=black,
  arc=3mm,
  boxrule=0.5pt,
  breakable,
  left=4mm,
  right=4mm,
  top=3mm,
  bottom=3mm,
  parbox=false,
  before skip=\medskipamount,
  after skip=\medskipamount,
  fontupper=\footnotesize 
}

\usepackage{graphicx}      
\usepackage{amsmath,amssymb,amsfonts} 
\usepackage{xcolor}        
\usepackage{cite}          

\usepackage{siunitx}       
\usepackage{fancyhdr}      

\usepackage{lipsum}        

\usepackage[scaled]{helvet}
\usepackage[T1]{fontenc}

\newcommand{\ISBN}{UNKNOWN ISBN}

\newcommand{\PaperTitle}{A title for your paper}
\newcommand{\PaperDate}{01 April 1900}
\newcommand{\AuthorFooter}{Gun, J. and Oksanen T.}
\newcommand{\PaperKeywords}{Kw1, Kw2, Kw3}

\makeatletter
\let\old@ssect\@ssect 
\makeatother

\usepackage[hidelinks]{hyperref}      
\hypersetup{hidelinks}
\usepackage{cleveref}      

\makeatletter
\def\@ssect#1#2#3#4#5#6{%
	\NR@gettitle{#6}
	\old@ssect{#1}{#2}{#3}{#4}{#5}{#6}
}
\makeatother

\renewcommand{\ISBN}{978-3-911430-11-1}
\renewcommand{\PaperTitle}{Agri-Query: A Case Study on RAG vs. Long-Context LLMs for Cross-Lingual Technical Question Answering}
\renewcommand{\PaperDate}{06 March 2026}
\renewcommand{\AuthorFooter}{Gun, J. and Oksanen T.}
\renewcommand{\PaperKeywords}{Agricultural Question Answering, Agricultural Machinery Manuals, Operator Support Systems, Industrial AI, Knowledge Extraction, cross-lingual information retrieval, Multilingual QA (Agriculture), long-document understanding, retrieval-augmented generation (RAG), large language models (LLMs)}

\hypersetup{
    pdfinfo={
        Title={\PaperTitle},
        Author={\AuthorFooter},
        Keywords={\PaperKeywords},
        Publisher={TUM Agrimechatronics},
        ISBN={\ISBN}
    }
}

\begin{document}

\setcitestyle{numbers} 
\begin{frontmatter}

   \title{\PaperTitle}

   \author[First]{Julius Gun,}
   \author[First]{Timo Oksanen}

   \address[First]{Technical University of Munich, Germany\\ Professorship of Agrimechatronics\\E-mail: timo.oksanen@tum.de\\ Munich Institute of Robotics and Machine Intelligence (MIRMI)}

   \begin{abstract}
	We present a case study evaluating large language models (LLMs) with 128K-token context windows on a technical question answering (QA) task. Our benchmark is built on a user manual for an agricultural machine, available in English, French, and German. It simulates a cross-lingual information retrieval scenario where questions are posed in English against all three language versions of the manual. The evaluation focuses on realistic "needle-in-a-haystack" challenges and includes unanswerable questions to test for hallucinations. We compare nine long-context LLMs using direct prompting against three Retrieval-Augmented Generation (RAG) strategies (keyword, semantic, hybrid), with an LLM-as-a-judge for evaluation. Our findings for this specific manual show that Hybrid RAG consistently outperforms direct long-context prompting. Models like Gemini 2.5 Flash and the smaller Qwen 2.5 7B achieve high accuracy (over 85\%) across all languages with RAG. This paper contributes a detailed analysis of LLM performance in a specialized industrial domain and an open framework for similar evaluations, highlighting practical trade-offs and challenges.
   \end{abstract}


\end{frontmatter}

\section{Introduction}
Technical user manuals are essential for all equipment. Modern European agricultural machinery is sophisticated with mechatronics and also highly regulated by the European Union for safety. These extensive documents provide comprehensive guidelines covering mechanical, electronic, and agronomical aspects.

Because Europe has many language areas, manufacturers must translate and maintain these manuals in multiple languages. This real-world scenario provides a practical basis for benchmarking the cross-lingual QA capabilities of LLMs. In this paper, we benchmark several state-of-the-art models to assess their QA robustness. For this benchmark, we curated a QA set based on domain expertise, focusing on critical operational and safety information, and including unanswerable questions to test the LLM's ability to avoid hallucinations.

The questions present a needle-in-a-haystack challenge where the answer is typically found in a single location within the user manual. This paper presents a case study comparing RAG approaches against a direct long-context method across various models and languages. 

\section{Related Work}
The ability of LLMs to understand long documents is an active research area. While Retrieval-Augmented Generation (RAG) can outperform Long-Context (LC) models \citep{yu2024defenserageralongcontext}, the performance is inconsistent across different tasks and datasets \citep{wang2024leavedocumentbehindbenchmarking, li2025larabenchmarkingretrievalaugmentedgeneration}. Newer LC models with large context windows show strong performance without RAG. However, RAG systems are often more resource-efficient and cheaper to maintain than LC systems \citep{li2024retrievalaugmentedgenerationlongcontext}.

\begin{figure}[H] 
  \includegraphics[width=\columnwidth]{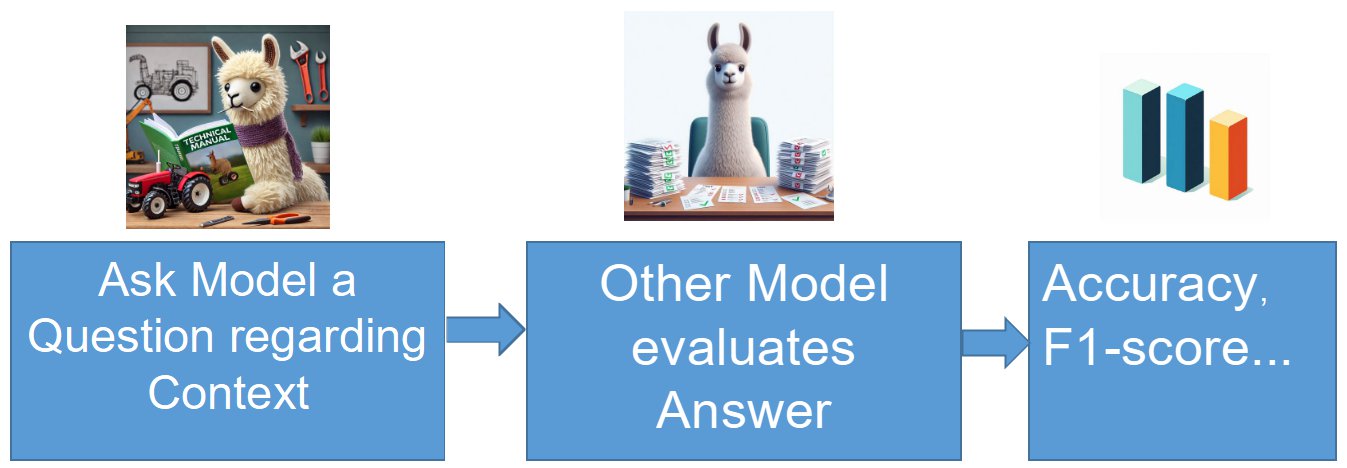}
  \caption{Process of data acquisition {\tiny (Illustrations created by the authors using Microsoft Bing Image Creator powered by DALL-E 3).}}
  \label{fig:simple_process}
\end{figure}
\newpage
\section{Materials}

\subsection{Models / LLM}
We tested several openly available LLMs, detailed in Table \ref{tab:model_overview_std}, and the proprietary Gemini 2.5 Flash model. All models used a temperature of 0 for deterministic outputs. We used an LLM-as-a-judge framework for automatic evaluation, comparing model outputs to ground-truth answers. Gemma 3 evaluated RAG results, while Gemma 2 evaluated long-context results. We acknowledge that using different, unvalidated judge models is a limitation of this study.

\begin{table}[htbp] 
    \centering 
    \caption{Overview of LLMs used in this study}
      \tiny 

    \label{tab:model_overview_std} 
    \begin{tabular}{lllr} 
        \toprule 
        \textbf{Model} & \textbf{Size} & \textbf{Context} & \textbf{Quantiz.} \\ 
        \midrule 
        Qwen 3 & 8B & 128k & Q4\_K\_M\\
        Qwen 2.5    & 7B    & 128k & Q4\_K\_M \\
        DeepSeek-R1 & 1.5B  & 128k & Q4\_K\_M \\ 
        DeepSeek-R1 & 8B    & 128k & Q4\_K\_M \\
        Gemma 2 (eval.) & 9B & 8k & Q4\_0 \\
        Gemma 3 (eval.) & 12B & 128k & Q4\_K\_M   \\
        Phi-3 Medium      & 14B & 128k & Q4\_0 \\
        Llama 3.1   & 8B    & 128k & Q4\_K\_M \\
        Llama 3.2   & 1B    & 128k & Q8\_0 \\
        Llama 3.2   & 3B    & 128k & Q4\_K\_M \\
        Gemini 2.5 Flash & - & 1M & -\\
        \bottomrule 
    \end{tabular}
\end{table}

\subsection{User Manual}
Our test case is the user manual for the Kverneland Exacta-TLX Geospread GS3, a mechatronic fertilizer spreader. We chose this manual because our familiarity with the machine aided in creating the QA set. We used the official English \citep{KvernelandManualEN}, French \citep{KvernelandManualFR}, and German \citep{KvernelandManualDE} versions. Each 165-page manual contains approximately 59k tokens and has an identical layout across languages, ensuring consistent page numbering for cross-lingual tests.

\section{Methods}
This work involved document preparation, QA dataset creation, long-context testing, and RAG system implementation. We converted the PDF manuals to Markdown format using the Docling library \citep{Docling}. We developed a small wrapper around the library to enable page-wise conversion. All experiments ran on a single NVIDIA RTX 6000 GPU, totaling approximately 80 GPU hours.

We created a QA test set of 108 questions from our domain expertise, focusing on critical operational and safety information. The dataset is balanced with 54 answerable and 54 unanswerable questions to test for hallucinations. To isolate cross-lingual retrieval capabilities, all questions were posed in English, following benchmarks like XTREME \citep{hu2020xtrememassivelymultilingualmultitask}. Appendix \ref{sec:appendix_qa_dataset} shows example questions.

Figure \ref{fig:simple_process} illustrates our evaluation process. First, a relevant context is selected. Second, an LLM is prompted with a question about the context. Third, an evaluator LLM assesses the answer's correctness.

\subsection{RAG system}
We tested three RAG retrieval methods. For all methods, the document was split into chunks of 200 tokens with a 100-token overlap. We used the embedding model \texttt{gte-Qwen2-7B-instruct} \citep{li2023} for semantic and hybrid retrieval. This model was chosen for its strong performance on the Massive Multilingual Text Embedding Benchmark (MTEB) \citep{enevoldsen2025mmtebmassivemultilingualtext}, making it well-suited for our cross-lingual tests. For each question, we retrieved the top three most relevant chunks. These RAG hyperparameters were fixed for all experiments; a sensitivity analysis is a subject for future work. We then provided these chunks and the question to the LLM. The prompts are available in Appendix \ref{sec:appendix_prompts}.
\subsubsection{Keyword-based Retrieval}
The BM25 algorithm \citep{BM25} was used for keyword-based retrieval. This method ranks documents based on the frequency of query terms within them, adjusted for document length and term rarity across the corpus. While efficient, it can fail if queries use synonyms not present in the text.
\subsubsection{Semantic Retrieval}
This method finds relevant chunks based on semantic meaning. Text is converted into numerical vectors (embeddings). The manual's chunks were vectorized and stored in a local ChromaDB vector database. During a query, the input question is also vectorized, and the database is searched for chunks with the highest cosine similarity to the query vector. Semantic retrieval may sometimes miss important query words. This method is more computationally intensive than keyword retrieval but can find relevant results even if phrasing differs from the document.

\subsubsection{Hybrid Retrieval}
Hybrid retrieval combines keyword-based (BM25) and semantic retrieval. This approach leverages the strengths of both methods. We perform keyword and semantic searches independently and then merge their ranked results using Reciprocal Rank Fusion (RRF) \citep{RRF} to produce a final, more robust ranking. RRF computes a new score for each retrieved chunk by summing the inverse of its rank from each retrieval list. This method effectively prioritizes chunks that consistently rank high across different search strategies, mitigating the weaknesses of any single method.

\subsection{Long-Context Testing}
We assessed direct long-context capabilities by providing LLMs with context and a question, without RAG or fine-tuning. For answerable questions, the context included the target page containing the answer. For unanswerable questions, a thematically related page was used. We simulated different context sizes (1k to 59k tokens) by adding surrounding pages as noise while preserving the original document order. This method tests the models' inherent understanding across various context lengths. Appendix \ref{sec:appendix_prompts} contains the prompts.

\section{Results}
We evaluated model performance using accuracy, F1 score, precision, recall, and specificity. In our evaluation, a \textbf{positive case} corresponds to an answerable question, and a \textbf{negative case} corresponds to an unanswerable question. This allows us to assess not only correctness but also the models' ability to avoid hallucination. Appendix \ref{sec:appendix_detailed_results} provides the formulas for these metrics.
\subsection{Long-Context QA Performance}
\begin{figure}[H]
    \centering
    \includegraphics[width=\columnwidth]{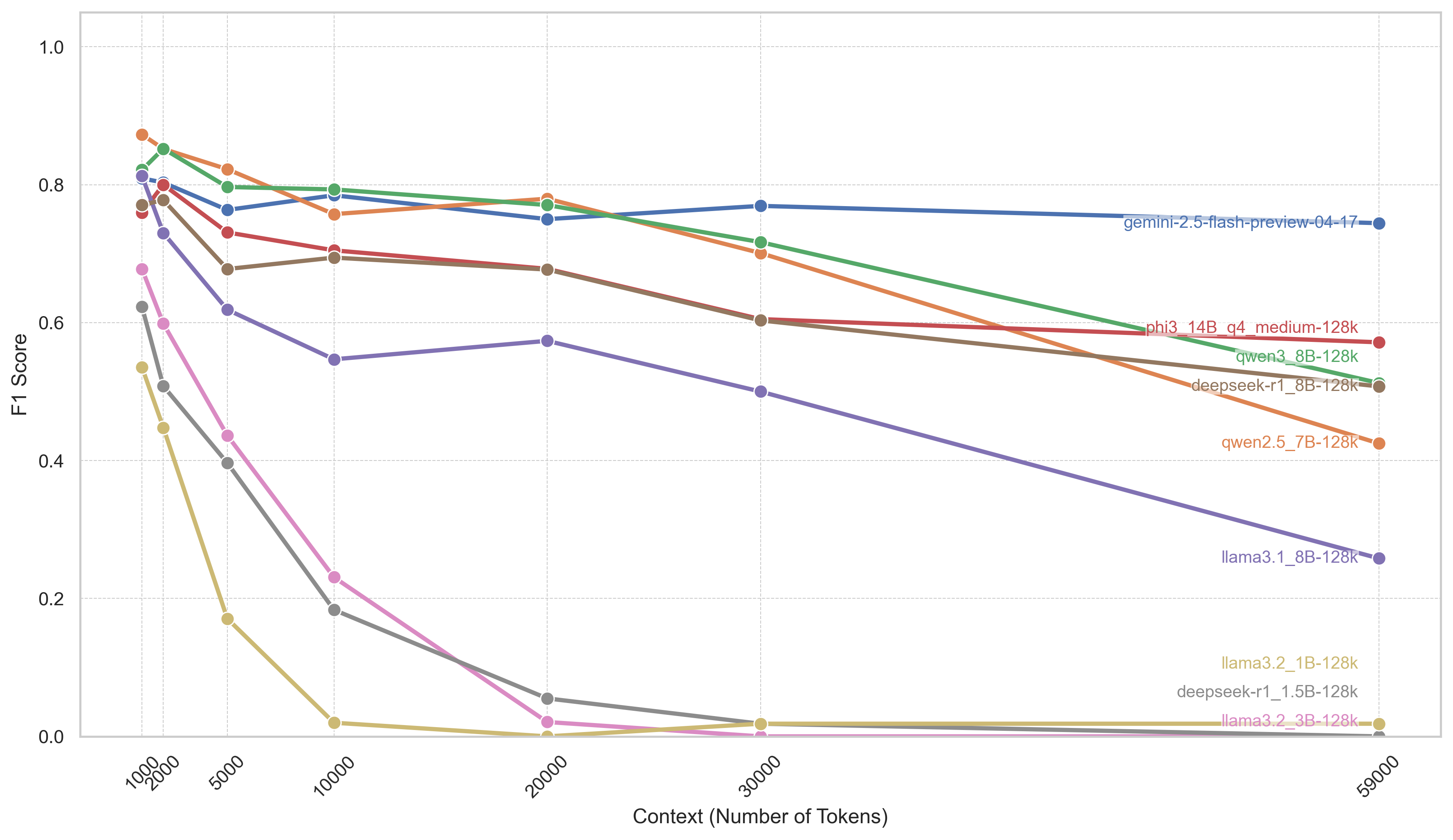}
    \caption{Long-Context QA: F1 score vs. noise: English.}
    \label{fig:zs_f1_noise_en_results}
\end{figure}

In our long-context tests without RAG, we provided the LLM with the relevant page plus surrounding pages as noise to reach specific token counts. Figure \ref{fig:zs_f1_noise_en_results} shows the F1 scores. The Lost in the Middle effect \citep{liu2023lostmiddlelanguagemodels} was pronounced for smaller models. Larger models like Gemini 2.5 Flash and Phi-3 14B performed better with the full manual context, though some performance degradation was still evident.

\subsection{RAG Performance}
\begin{figure}[H]  
    \centering
    \includegraphics[width=1\columnwidth]{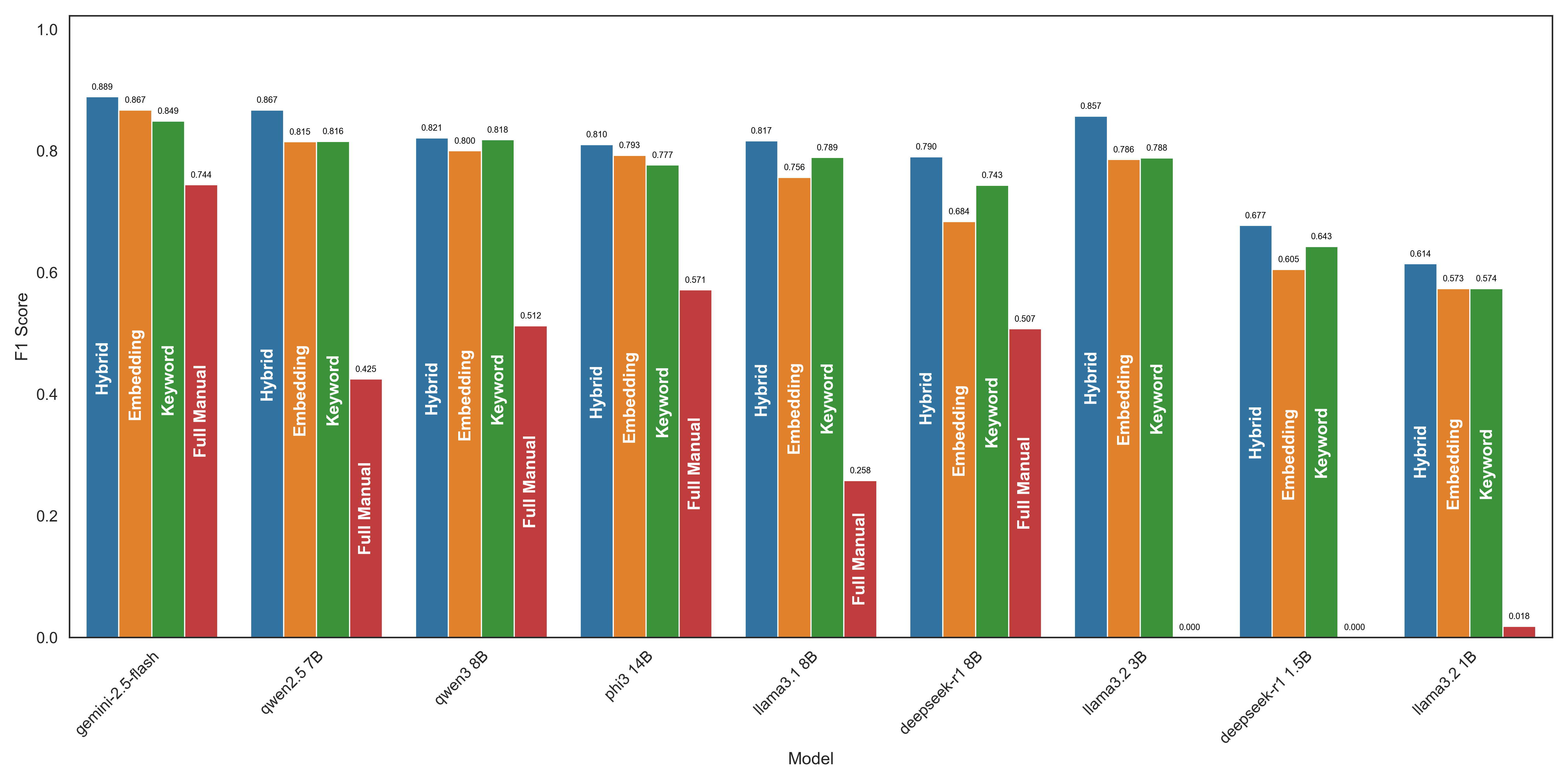} 
    \caption{F1 comparison for English language across RAG retrieval and Full Manual (59k tokens).}
    \label{fig:english_retrieval_f1_score_comparison}
\end{figure}

Next, RAG performance was tested with the same QA set and prompt structure. Figure \ref{fig:english_retrieval_f1_score_comparison} shows the F1 score. Hybrid retrieval consistently achieved the highest accuracy and F1 scores. Gemini 2.5 Flash had the highest accuracy, while naively inserting the full manual yielded significantly worse results. This indicates RAG, especially Hybrid retrieval, is better for needle-in-a-haystack tasks and suits technical documents like our agricultural manuals.
Notably,smaller models like Llama 3.2 3B and Qwen 2.5 7B also achieved high performance (Accuracy > 0.85, Specificity = 0.815), demonstrating that RAG can enable effective results on resource-constrained hardware.

Table \ref{tab:english_hybrid_performance_results} shows more detailed results for Hybrid RAG. Precision and recall are very similar across models. However, specificity is much lower for smaller models. This indicates smaller models are more likely to produce false positives (i.e., hallucinate answers to unanswerable questions), while larger models are more likely to produce false negatives (i.e., fail to find an existing answer).

\begin{table}[!htbp]
\caption{Performance on English Manual using Hybrid retrieval}
\label{tab:english_hybrid_performance_results}
  \tiny 
\centering 
\begin{tabular}{lccccc}
\toprule
Metric & Acc. & F1 & Prec. & Rec. & Spec. \\
LLM &  &  &  &  &  \\
\midrule
\midrule
Gemini 2.5 Flash & \textbf{0.880} & \textbf{0.889} & 0.825 & \textbf{0.963} & 0.796 \\
Qwen 2.5 7B & 0.861 & 0.867 & \textbf{0.831} & 0.907 & \textbf{0.815} \\
Qwen3 8B & 0.815 & 0.821 & 0.793 & 0.852 & 0.778 \\
Phi3 14B & 0.796 & 0.810 & 0.758 & 0.870 & 0.722 \\
Llama3.1 8B & 0.796 & 0.817 & 0.742 & 0.907 & 0.685 \\
Deepseek-R1 8B & 0.759 & 0.790 & 0.700 & 0.907 & 0.611 \\
Llama3.2 3B & 0.852 & 0.857 & 0.828 & 0.889 & \textbf{0.815} \\
Deepseek-R1 1.5B & 0.630 & 0.677 & 0.600 & 0.778 & 0.481 \\
Llama3.2 1B & 0.500 & 0.614 & 0.500 & 0.796 & 0.204 \\
\bottomrule
\end{tabular}
\end{table}
\subsection{Cross-lingual Performance using Hybrid RAG}
\begin{figure}[H]  
    \centering
    \includegraphics[width=0.8\columnwidth]{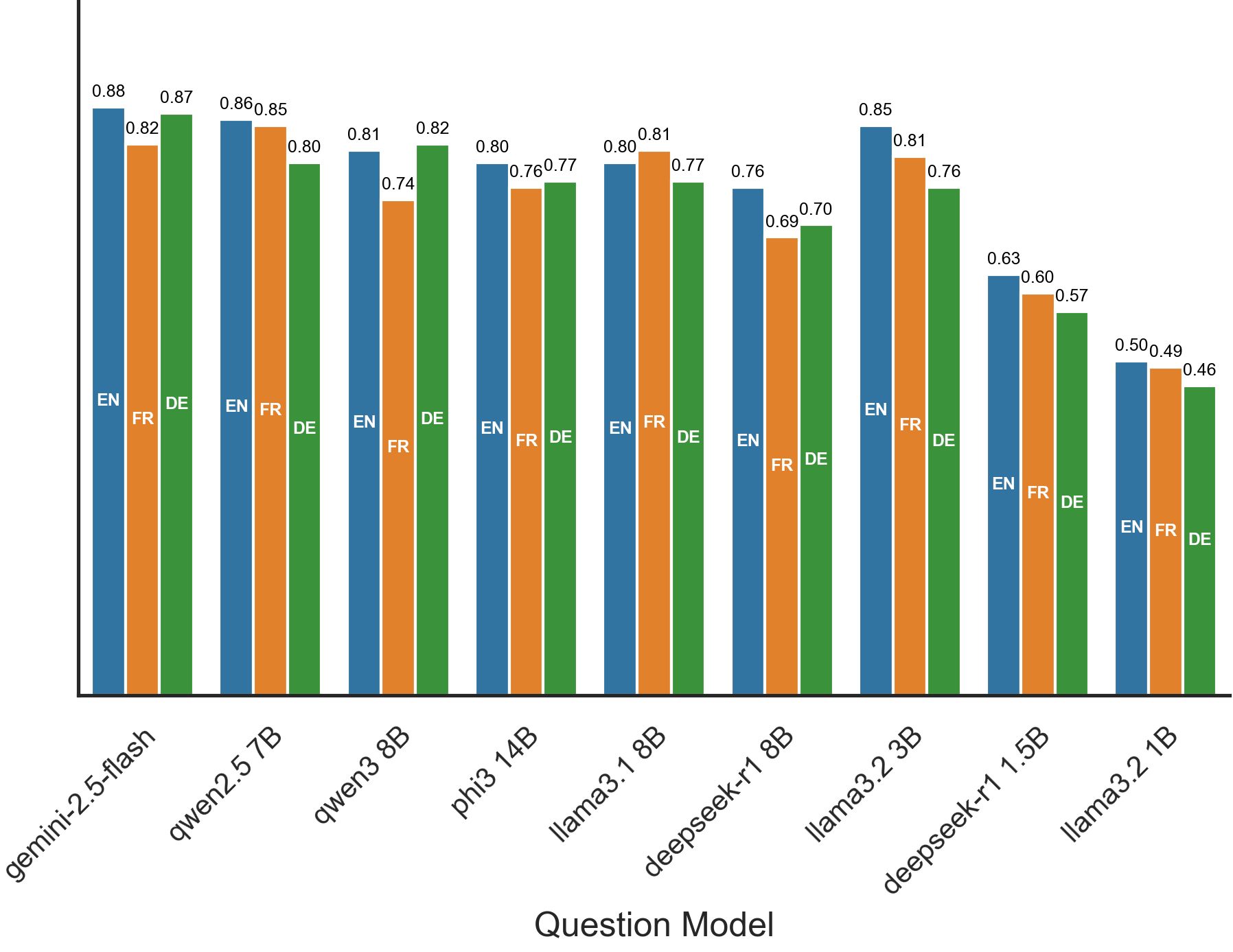}
    \caption{Accuracy comparison across different languages.}
    \label{fig:model_perf_acc_comp}
\end{figure}

\begin{figure}[H] 
    \centering
    \includegraphics[width=0.8\columnwidth]{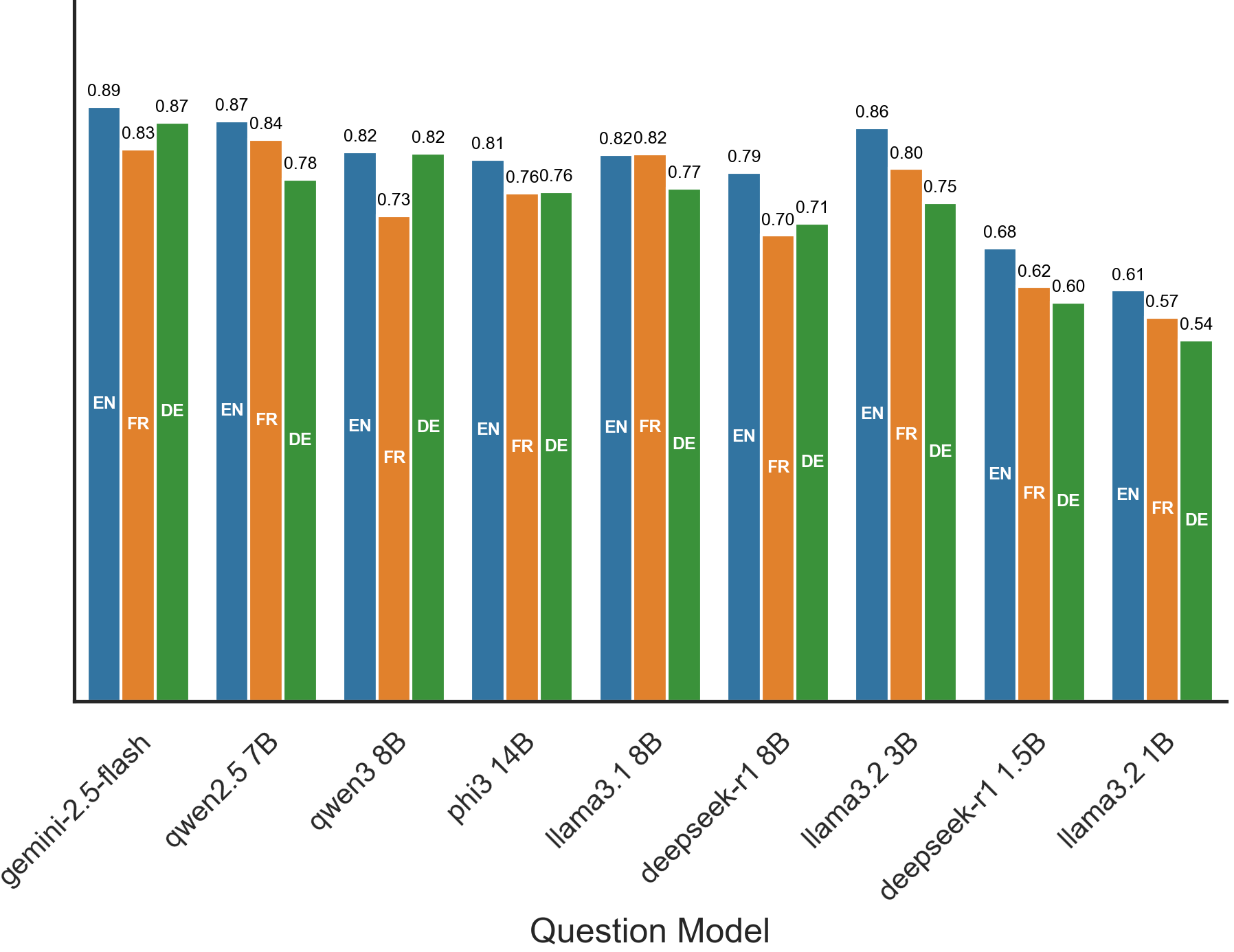}
    \caption{F1 score comparison across different languages.}
    \label{fig:model_perf_f1_comp}
\end{figure}

Lastly, we evaluated the models' cross-lingual information retrieval capabilities. As described, this setup involves posing questions in English against non-English documents (French and German) to assess the system's ability to bridge this language gap. Figure \ref{fig:model_perf_acc_comp} shows accuracy, and Figure \ref{fig:model_perf_f1_comp} shows the F1 score across English (EN), French (FR), and German (DE) for Hybrid RAG. For most models, performance on French or German was comparable to English, demonstrating that hybrid RAG with a strong multilingual embedding model offers robust cross-lingual retrieval.

\section{Discussion}
Our results offer a case study on applying LLMs to a real-world technical QA task, highlighting key points about RAG and long-context models in a specialized, multilingual domain.

\subsection{RAG vs. Long-Context}
For the tested agricultural manual, Hybrid RAG consistently outperformed the direct long-context approach. This was true even when comparing smaller models using RAG against larger models using the full context. The pronounced "Lost in the Middle" effect in our long-context tests (Figure \ref{fig:zs_f1_noise_en_results}) underscores the ongoing challenges large context models face in reliably locating specific facts within long, noisy inputs. Our results suggest that for applications requiring high-precision factual retrieval from dense technical documents, a well-configured RAG system remains a more robust choice.

\textbf{Cross-Lingual Capabilities:} The hybrid RAG approach demonstrated strong cross-lingual performance. High-performing models like Gemini 2.5 Flash and Qwen 2.5 7B maintained high accuracy when querying in English against French and German manuals. This indicates that the combination of a powerful multilingual embedding model and a capable LLM can effectively bridge language gaps for information retrieval tasks.

\textbf{Failure Modes:} A qualitative review revealed two primary failure modes: retrieval failure and hallucination. Retrieval failure occurs when the RAG system does not retrieve the correct context, a problem less frequent with the hybrid approach. Hallucination was more common, especially for unanswerable questions, with smaller models showing a higher tendency to invent answers (lower specificity in Table \ref{tab:english_hybrid_performance_results}).

\textbf{Potential Risks}
While LLMs can enhance access to information, they also pose risks of misinformation and over-reliance on AI-generated content. Users must exercise caution, especially with unanswerable questions, as models may generate plausible but incorrect answers.

\textbf{Future Research:} Future work should expand the benchmark to other technical domains to test generalizability. It could also explore more complex queries requiring information synthesis and test queries in the document's native language. A sensitivity analysis of RAG hyperparameters (e.g., chunk size, embedding model) is needed. 

\section{Conclusions}
We present a framework for benchmarking LLMs on RAG and long-context tasks. Our findings show that for the agricultural manual tested, RAG is highly effective, enabling even small models to achieve strong results. The Lost in the Middle effect highlights that context window length is a critical factor for long-context models. Finally, our results demonstrate that robust cross-lingual performance is achievable with Hybrid RAG paired with capable LLMs.

\section*{Limitations}
This study has several limitations.

\textbf{Scope and Generalizability:} The benchmark uses a single agricultural manual. Findings may not generalize to other domains or document types.

\textbf{Dataset:} The QA dataset is limited to 108 questions curated with domain expertise. A larger, more diverse dataset would provide greater statistical power. We did not perform statistical significance testing.

\textbf{Evaluation Methodology:} Our evaluation uses an LLM-as-a-judge framework not validated against human annotators, which may introduce bias. Using different judge models for RAG (Gemma 3) and long-context (Gemma 2) experiments is a confounding variable that complicates direct comparison.

\textbf{Experimental Design:}
\begin{itemize}
    \item \textbf{RAG Hyperparameters:} The RAG configuration was fixed (chunk size: 200, overlap: 100, top-k: 3) and used a single embedding model. The reported superiority of Hybrid RAG may be configuration-specific, and a hyperparameter sweep could yield different results.
    \item \textbf{Cross-Lingual Task:} Questions were posed only in English. This setup tests cross-lingual information retrieval but does not fully represent a scenario where a native speaker would query the document in their own language.
    \item \textbf{Question Complexity:} The questions primarily target factual, localized information. The benchmark does not assess the models' ability to synthesize information across multiple sections, reason about complex procedures, or interpret tables and figures.
\end{itemize}

\textbf{Reproducibility:} Only a single test run was conducted for each experiment, so we do not report variance in the results.

\section*{Declaration of generative AI and AI-assisted technologies in the manuscript preparation process}
AI tools were used by the authors to generate Figure \ref{fig:simple_process} (Microsoft Bing Image Creator powered by DALL-E 3). Gemini 2.5 Pro was also used for paraphrasing.

\section*{Code}
The code for the framework is available at \url{https://github.com/julius-gun/agriquery}.

\bibliographystyle{IEEEtranN} 
\bibliography{bibtex/literature}             

\newpage

\appendix

\section{QA Test Dataset}
\label{sec:appendix}
\label{sec:appendix_qa_dataset}

\subsection{Answerable question samples}

\begin{itemize}
    \item Q: What torque (in Nm) should be applied to the vane lock nuts? A: 50 Nm
    \item Q: What is the required grease level in mm below the filler opening for spreading disc gearboxes after the machine has stood still? A: 35 mm
    \item Q: Where is the main switch button to turn the control box on or off located? A: The main switch button is located on the upper left in the red extension.
    \item Q: How to enable fine application for dosing low application rate? A: Move the fine application handle to the fine dosing position on both sides.
    \item Q: Where is the RS 232 connector located? A: At the back of the control box.
    \item Q: How often should the agitator axle seal be replaced? A: Every season and after every 100 operational hours.
    \item Q: From which machine point is the spreading height measured to the ground or the crop? A: Measured from the bottom of the vanes.
    \item Q: What materials are required to perform the tray test? A: A measuring tape or ruler, a spirit level, 7 troughs, 7 graduated tubes, a funnel, a notebook, pen, calculator, this manual, and the software's instruction manual.
    \item Q: How many parts does the distribution meter have? A: Seven. 
    \item Q: Should the parking brake of the tractor be engaged before connecting the machine? A: Yes. 
    \item Q: How long should the main switch button be pressed to turn the control box on or off? A: At least 1 second. 
    \item Q: What is the overlap percentage for the full field spreading pattern? A: 100\% overlap. 
    \item Q: What determines the machine's working width? A: Spreading disc RPM
    \item Q: When shortening a coupling shaft, how far must profiled tubes at least overlap in mm? A: 150mm. 
\end{itemize}

\subsection{Unanswerable question samples}
\begin{itemize}
    \item Q: How much extra diesel does the tractor consume to use the Exacta-TLX GEOSPREAD? A: Not found in context
    \item Q: Is one-sided boundary spreading suitable for small gardens? A: Not found in context
    \item Q: Can IsoMatch Tellus be connected to an external mouse? A: Not found in context
    \item Q: Can the linkage pin for the tractor be made out of aluminium? A: Not found in context
    \item Q: What is the maximum height the fertilizer flies when spreading without GEOCONTROL headland? A: Not found in context
    \item Q: What happens if the machine grease nipples are never lubricated? A: Not found in context
    \item Q: What kind of protective safety gloves are needed for cleaning fertiliser remnants from the Exacta-TLX GEOSPREAD before welding? A: Not found in context
    \item Q: What specific 'grease' type is recommended for 'profiled tubes' of the  coupling shaft? A: Not found in context
    \item Q: What is the minimum 'baud rate' for 'RS 232 connection'? A: Not found in context
    \item Q: What is the recommended tire pressure 'range' for transport mode? A: Not found in context
    \item Q: What is the drain rate in liters per minute of 'drain kit' for hopper emptying? A: Not found in context
    \item Q: How many hours of continuous operation can the IsoMatch Tellus operate before flattening a typical tractor battery if left switched on with the engine off? A: Not found in context
    \item Q: What specific paint should be used to paint any damaged paintwork at the end of the season the machine in preparation for winter storage? A: Not found in context
    \item Q: Can the IsoMatch universal ISOBUS terminal be used to check the weather? A: Not found in context
\end{itemize}

\onecolumn

\section{Prompts}
\label{sec:appendix_prompts}

\subsection{Question prompts}
\begin{appendixpromptstyle}
\texttt{<purpose>} Extract a precise, concise answer to the question from the given context. Adhere strictly to the instructions. Base your answer on the context. \texttt{</purpose>}\par
\texttt{<instructions>}\par
    \hspace*{1em}\texttt{<instruction>} Read the entire context carefully \texttt{</instruction>}\par
    \hspace*{1em}\texttt{<instruction>} Focus ONLY on the specific information related to the question \texttt{</instruction>}\par
    \hspace*{1em}\texttt{<instruction>} Provide an extremely precise answer \texttt{</instruction>}\par
    \hspace*{1em}\texttt{<instruction>} Match the expected answer format exactly \texttt{</instruction>}\par
    \hspace*{1em}\texttt{<instruction>} If unsure, respond with "Unknown" or "Not found in context" \texttt{</instruction>}\par
	\hspace*{1em}\texttt{<instruction>} Answer in English \texttt{</instruction>}\par
\texttt{</instructions>}\par
\vspace{1ex} 

\texttt{<context>}\par
    \hspace*{1em}\highlightblue{{context}}\par
\texttt{</context>}\par
\texttt{<question>}\par
    \hspace*{1em}\highlightblue{{question}}\par
\texttt{</question>}\par
\texttt{<answer>}\par
    \hspace*{1em}[Carefully extract the EXACT information that directly answers the question, keeping it as brief and precise as possible]\par
\texttt{</answer>}
\end{appendixpromptstyle}

\subsection{Evaluation prompt}
\begin{appendixpromptstyle}
\texttt{<purpose>} ANSWER COMPARISON TASK. Do ANSWER\_ONE and ANSWER\_TWO convey the same information regarding the QUESTION? Adhere strictly to the INSTRUCTIONS. Base your ANSWER on the CONTEXT. \texttt{</purpose>}\par
\texttt{<INSTRUCTIONS>}\par
    \hspace*{1em}\texttt{<instruction>} - Respond 'yes' if ANSWER\_ONE and the ANSWER\_TWO convey the SAME TECHNICAL MEANING \texttt{</instruction>}\par
    \hspace*{1em}\texttt{<instruction>} - Consider 'yes' if differences are INSIGNIFICANT to the core technical content \texttt{</instruction>}\par
    \hspace*{1em}\texttt{<instruction>} - Respond 'no' ONLY if there are MEANINGFUL differences that alter the technical understanding \texttt{</instruction>}\par
    \hspace*{1em}\texttt{<instruction>} - Assess the SUBSTANCE of the information, not surface-level variations \texttt{</instruction>}\par
    \hspace*{1em}\texttt{<instruction>} - Answer ONLY with yes or no \texttt{</instruction>}\par
    \hspace*{1em}\texttt{<instruction>} - Don't provide additional information \texttt{</instruction>}\par
\texttt{</INSTRUCTIONS>}\par
\vspace{1ex}

\texttt{<CONTEXT>}\par
    \hspace*{1em}\texttt{<QUESTION>}\par \highlightblue{{question}}\par \texttt{</QUESTION>}\par
    \hspace*{1em}\texttt{<ANSWER\_ONE>}\par \highlightblue{{model_answer}}\par \texttt{</ANSWER\_ONE>}\par
    \hspace*{1em}\texttt{<ANSWER\_TWO>}\par \highlightblue{{expected_answer}}\par \texttt{</ANSWER\_TWO>}\par
\texttt{</CONTEXT>}\par
\vspace{1ex}

\texttt{<ANSWER>}\par
    \hspace*{1em}(yes/no)\par
\texttt{</ANSWER>}
\end{appendixpromptstyle}
\clearpage
\twocolumn
\section{Detailed Results}
\label{sec:appendix_detailed_results}

We used the following metrics to evaluate the performance of the models. For a given question, the outcome is classified into one of four categories based on whether the question is answerable and whether the model's response is correct. A positive case is an answerable question, and a negative case is an unanswerable question.
\begin{itemize}
    \item \textbf{TP (True Positive):} The model correctly answers an answerable question.
    \item \textbf{TN (True Negative):} The model correctly identifies an unanswerable question (e.g., by responding "Not found in context").
    \item \textbf{FP (False Positive):} The model provides an incorrect answer to an unanswerable question (hallucination). 
    \item \textbf{FN (False Negative):} The model fails to answer an answerable question correctly. 
\end{itemize}

\begin{equation*}
\begin{aligned}
\text{Accuracy} &= \frac{TP + TN}{TP + TN + FP + FN} \\
\text{Precision} &= \frac{TP}{TP + FP} \\
\text{Recall (Sensitivity)} &= \frac{TP}{TP + FN} \\
\text{Specificity} &= \frac{TN}{TN + FP} \\
F_1 \text{ Score} &= \quad 2 \cdot \frac{\text{Precision} \cdot \text{Recall}}{\text{Precision} + \text{Recall}}
\end{aligned}
\end{equation*}

\begin{figure}[H]  
    \centering
    \includegraphics[width=1\columnwidth]{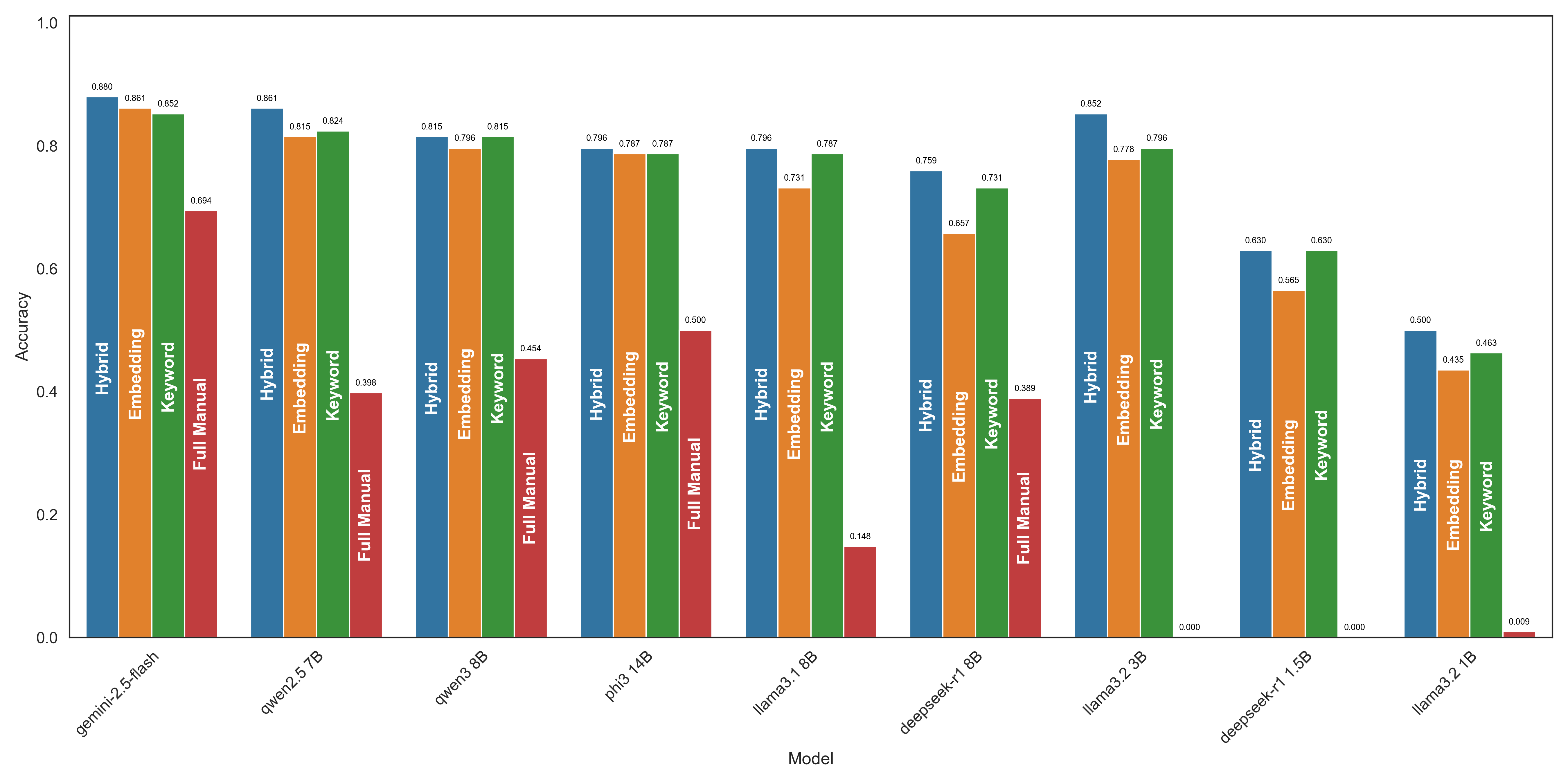} 
    \caption{Accuracy comparison for English language across RAG retrieval and Full Manual (59k tokens).}
    \label{fig:english_retrieval_accuracy_comparison}
\end{figure}

\subsection{Keyword RAG Performance}
\label{sec:keyword_rag}
This section includes tables detailing the performance metrics for models utilizing the Keyword RAG retrieval algorithm across various languages.
\begin{table}[!htbp]
\caption{Performance of English Keyword Models}
\label{tab:english_keyword_performance}
  \tiny
\begin{tabular}{lccccc}
\toprule
Metric & Acc. & F1 & Prec. & Rec. & Spec. \\
LLM &  &  &  &  &  \\
\midrule
\midrule
Gemini 2.5 Flash & \textbf{0.852} & \textbf{0.849} & \textbf{0.865} & \textbf{0.833} & \textbf{0.870} \\
Qwen 2.5 7B & 0.824 & 0.816 & 0.857 & 0.778 & \textbf{0.870} \\
Qwen3 8B & 0.815 & 0.818 & 0.804 & \textbf{0.833} & 0.796 \\
Phi3 14B & 0.787 & 0.777 & 0.816 & 0.741 & 0.833 \\
Llama3.1 8B & 0.787 & 0.789 & 0.782 & 0.796 & 0.778 \\
Deepseek-R1 8B & 0.731 & 0.743 & 0.712 & 0.778 & 0.685 \\
Llama3.2 3B & 0.796 & 0.788 & 0.820 & 0.759 & 0.833 \\
Deepseek-R1 1.5B & 0.630 & 0.643 & 0.621 & 0.667 & 0.593 \\
Llama3.2 1B & 0.463 & 0.574 & 0.476 & 0.722 & 0.204 \\
\bottomrule
\end{tabular}
\end{table}

\begin{table}[!htbp]
\caption{Performance of French Keyword Models}
\label{tab:french_keyword_performance}
  \tiny
\begin{tabular}{lccccc}
\toprule
Metric & Acc. & F1 & Prec. & Rec. & Spec. \\
LLM &  &  &  &  &  \\
\midrule
\midrule
Gemini 2.5 Flash & \textbf{0.583} & 0.328 & 0.846 & 0.204 & 0.963 \\
Qwen 2.5 7B & 0.556 & 0.273 & 0.750 & 0.167 & 0.944 \\
Qwen3 8B & 0.556 & 0.294 & 0.714 & 0.185 & 0.926 \\
Phi3 14B & 0.546 & 0.310 & 0.647 & 0.204 & 0.889 \\
Llama3.1 8B & \textbf{0.583} & \textbf{0.348} & 0.800 & \textbf{0.222} & 0.944 \\
Deepseek-R1 8B & 0.491 & 0.267 & 0.476 & 0.185 & 0.796 \\
Llama3.2 3B & 0.565 & 0.230 & \textbf{1.000} & 0.130 & \textbf{1.000} \\
Deepseek-R1 1.5B & 0.389 & 0.233 & 0.312 & 0.185 & 0.593 \\
Llama3.2 1B & 0.213 & 0.206 & 0.208 & 0.204 & 0.222 \\
\bottomrule
\end{tabular}
\end{table}

\begin{table}[!htbp]
\caption{Performance of German Keyword Models}
\label{tab:german_keyword_performance}
  \tiny
\begin{tabular}{lccccc}
\toprule
Metric & Acc. & F1 & Prec. & Rec. & Spec. \\
LLM &  &  &  &  &  \\
\midrule
\midrule
Gemini 2.5 Flash & 0.528 & 0.190 & 0.667 & 0.111 & 0.944 \\
Qwen 2.5 7B & \textbf{0.537} & 0.138 & \textbf{1.000} & 0.074 & \textbf{1.000} \\
Qwen3 8B & \textbf{0.537} & 0.194 & 0.750 & 0.111 & 0.963 \\
Phi3 14B & 0.528 & 0.164 & 0.714 & 0.093 & 0.963 \\
Llama3.1 8B & 0.519 & 0.161 & 0.625 & 0.093 & 0.944 \\
Deepseek-R1 8B & 0.472 & 0.174 & 0.400 & 0.111 & 0.833 \\
Llama3.2 3B & 0.509 & 0.102 & 0.600 & 0.056 & 0.963 \\
Deepseek-R1 1.5B & 0.306 & 0.096 & 0.138 & 0.074 & 0.537 \\
Llama3.2 1B & 0.398 & \textbf{0.198} & 0.296 & \textbf{0.148} & 0.648 \\
\bottomrule
\end{tabular}
\end{table}

\subsection{Embedding RAG Performance}
\label{sec:embedding_rag}
This section includes tables detailing the performance metrics for models utilizing the Embedding RAG retrieval algorithm across various languages.
\begin{table}[!htbp]
\caption{Performance of English Embedding Models}
\label{tab:english_embedding_performance}
  \tiny
\begin{tabular}{lccccc}
\toprule
Metric & Acc. & F1 & Prec. & Rec. & Spec. \\
LLM &  &  &  &  &  \\
\midrule
\midrule
Gemini 2.5 Flash & \textbf{0.861} & \textbf{0.867} & \textbf{0.831} & \textbf{0.907} & \textbf{0.815} \\
Qwen 2.5 7B & 0.815 & 0.815 & 0.815 & 0.815 & \textbf{0.815} \\
Qwen3 8B & 0.796 & 0.800 & 0.786 & 0.815 & 0.778 \\
Phi3 14B & 0.787 & 0.793 & 0.772 & 0.815 & 0.759 \\
Llama3.1 8B & 0.731 & 0.756 & 0.692 & 0.833 & 0.630 \\
Deepseek-R1 8B & 0.657 & 0.684 & 0.635 & 0.741 & 0.574 \\
Llama3.2 3B & 0.778 & 0.786 & 0.759 & 0.815 & 0.741 \\
Deepseek-R1 1.5B & 0.565 & 0.605 & 0.554 & 0.667 & 0.463 \\
Llama3.2 1B & 0.435 & 0.573 & 0.461 & 0.759 & 0.111 \\
\bottomrule
\end{tabular}
\end{table}

\begin{table}[!htbp]
\caption{Performance of French Embedding Models}
\label{tab:french_embedding_performance}
  \tiny
\begin{tabular}{lccccc}
\toprule
Metric & Acc. & F1 & Prec. & Rec. & Spec. \\
LLM &  &  &  &  &  \\
\midrule
\midrule
Gemini 2.5 Flash & \textbf{0.806} & \textbf{0.784} & \textbf{0.884} & \textbf{0.704} & \textbf{0.907} \\
Qwen 2.5 7B & 0.741 & 0.714 & 0.795 & 0.648 & 0.833 \\
Qwen3 8B & 0.657 & 0.626 & 0.689 & 0.574 & 0.741 \\
Phi3 14B & 0.648 & 0.642 & 0.654 & 0.630 & 0.667 \\
Llama3.1 8B & 0.704 & 0.704 & 0.704 & \textbf{0.704} & 0.704 \\
Deepseek-R1 8B & 0.537 & 0.545 & 0.536 & 0.556 & 0.519 \\
Llama3.2 3B & 0.648 & 0.548 & 0.767 & 0.426 & 0.870 \\
Deepseek-R1 1.5B & 0.380 & 0.385 & 0.382 & 0.389 & 0.370 \\
Llama3.2 1B & 0.324 & 0.425 & 0.370 & 0.500 & 0.148 \\
\bottomrule
\end{tabular}
\end{table}

\begin{table}[!htbp]
\caption{Performance of German Embedding Models}
\label{tab:german_embedding_performance}
  \tiny
\begin{tabular}{lccccc}
\toprule
Metric & Acc. & F1 & Prec. & Rec. & Spec. \\
LLM &  &  &  &  &  \\
\midrule
\midrule
Gemini 2.5 Flash & \textbf{0.824} & \textbf{0.822} & \textbf{0.830} & \textbf{0.815} & \textbf{0.833} \\
Qwen 2.5 7B & 0.759 & 0.740 & 0.804 & 0.685 & \textbf{0.833} \\
Qwen3 8B & 0.694 & 0.692 & 0.698 & 0.685 & 0.704 \\
Phi3 14B & 0.722 & 0.732 & 0.707 & 0.759 & 0.685 \\
Llama3.1 8B & 0.722 & 0.732 & 0.707 & 0.759 & 0.685 \\
Deepseek-R1 8B & 0.537 & 0.583 & 0.530 & 0.648 & 0.426 \\
Llama3.2 3B & 0.713 & 0.674 & 0.780 & 0.593 & \textbf{0.833} \\
Deepseek-R1 1.5B & 0.389 & 0.431 & 0.403 & 0.463 & 0.315 \\
Llama3.2 1B & 0.565 & 0.561 & 0.566 & 0.556 & 0.574 \\
\bottomrule
\end{tabular}
\end{table}

\subsection{Hybrid RAG Performance}
\label{sec:hybrid_rag}
This section includes tables detailing the performance metrics for models utilizing the Hybrid RAG retrieval algorithm across various languages.
\begin{table}[!htbp]
\caption{Performance of English Hybrid Models}
\label{tab:english_hybrid_performance}
  \tiny
\begin{tabular}{lccccc}
\toprule
Metric & Acc. & F1 & Prec. & Rec. & Spec. \\
LLM &  &  &  &  &  \\
\midrule
\midrule
Gemini 2.5 Flash & \textbf{0.880} & \textbf{0.889} & 0.825 & \textbf{0.963} & 0.796 \\
Qwen 2.5 7B & 0.861 & 0.867 & \textbf{0.831} & 0.907 & \textbf{0.815} \\
Qwen3 8B & 0.815 & 0.821 & 0.793 & 0.852 & 0.778 \\
Phi3 14B & 0.796 & 0.810 & 0.758 & 0.870 & 0.722 \\
Llama3.1 8B & 0.796 & 0.817 & 0.742 & 0.907 & 0.685 \\
Deepseek-R1 8B & 0.759 & 0.790 & 0.700 & 0.907 & 0.611 \\
Llama3.2 3B & 0.852 & 0.857 & 0.828 & 0.889 & \textbf{0.815} \\
Deepseek-R1 1.5B & 0.630 & 0.677 & 0.600 & 0.778 & 0.481 \\
Llama3.2 1B & 0.500 & 0.614 & 0.500 & 0.796 & 0.204 \\
\bottomrule
\end{tabular}
\end{table}

\begin{table}[!htbp]
\caption{Performance of French Hybrid Models}
\label{tab:french_hybrid_performance}
  \tiny
\begin{tabular}{lccccc}
\toprule
Metric & Acc. & F1 & Prec. & Rec. & Spec. \\
LLM &  &  &  &  &  \\
\midrule
\midrule
Gemini 2.5 Flash & 0.824 & 0.826 & 0.818 & \textbf{0.833} & 0.815 \\
Qwen 2.5 7B & \textbf{0.852} & \textbf{0.840} & \textbf{0.913} & 0.778 & \textbf{0.926} \\
Qwen3 8B & 0.741 & 0.725 & 0.771 & 0.685 & 0.796 \\
Phi3 14B & 0.759 & 0.759 & 0.759 & 0.759 & 0.759 \\
Llama3.1 8B & 0.815 & 0.818 & 0.804 & \textbf{0.833} & 0.796 \\
Deepseek-R1 8B & 0.685 & 0.696 & 0.672 & 0.722 & 0.648 \\
Llama3.2 3B & 0.806 & 0.796 & 0.837 & 0.759 & 0.852 \\
Deepseek-R1 1.5B & 0.602 & 0.619 & 0.593 & 0.648 & 0.556 \\
Llama3.2 1B & 0.491 & 0.574 & 0.493 & 0.685 & 0.296 \\
\bottomrule
\end{tabular}
\end{table}

\begin{table}[!htbp]
\caption{Performance of German Hybrid Models}
\label{tab:german_hybrid_performance}
  \tiny
\begin{tabular}{lccccc}
\toprule
Metric & Acc. & F1 & Prec. & Rec. & Spec. \\
LLM &  &  &  &  &  \\
\midrule
\midrule
Gemini 2.5 Flash & \textbf{0.870} & \textbf{0.865} & \textbf{0.900} & \textbf{0.833} & \textbf{0.907} \\
Qwen 2.5 7B & 0.796 & 0.780 & 0.848 & 0.722 & 0.870 \\
Qwen3 8B & 0.824 & 0.819 & 0.843 & 0.796 & 0.852 \\
Phi3 14B & 0.769 & 0.762 & 0.784 & 0.741 & 0.796 \\
Llama3.1 8B & 0.769 & 0.766 & 0.774 & 0.759 & 0.778 \\
Deepseek-R1 8B & 0.704 & 0.714 & 0.690 & 0.741 & 0.667 \\
Llama3.2 3B & 0.759 & 0.745 & 0.792 & 0.704 & 0.815 \\
Deepseek-R1 1.5B & 0.574 & 0.596 & 0.567 & 0.630 & 0.519 \\
Llama3.2 1B & 0.463 & 0.540 & 0.472 & 0.630 & 0.296 \\
\bottomrule
\end{tabular}
\end{table}

\subsection{Full Manual Performance (Long-Context @ approx. 59k Tokens)}
\label{sec:full_manual_performance}
This section presents performance metrics for models under the "Full Manual" configuration, corresponding to Long-Context evaluations with a context of approximately 59,000 tokens (entire document).
\begin{table}[!htbp]
\caption{Performance of English Full Manual Models}
\label{tab:english_full_manual_performance}
  \tiny
\begin{tabular}{lccccc}
\toprule
Metric & Acc. & F1 & Prec. & Rec. & Spec. \\
LLM &  &  &  &  &  \\
\midrule
\midrule
Gemini 2.5 Flash & \textbf{0.694} & \textbf{0.744} & \textbf{0.640} & \textbf{0.889} & \textbf{0.500} \\
Qwen 2.5 7B & 0.398 & 0.425 & 0.407 & 0.444 & 0.352 \\
Qwen3 8B & 0.454 & 0.512 & 0.463 & 0.574 & 0.333 \\
Phi3 14B & 0.500 & 0.571 & 0.500 & 0.667 & 0.333 \\
Llama3.1 8B & 0.148 & 0.258 & 0.229 & 0.296 & 0.000 \\
Deepseek-R1 8B & 0.389 & 0.507 & 0.425 & 0.630 & 0.148 \\
Llama3.2 3B & 0.000 & 0.000 & 0.000 & 0.000 & 0.000 \\
Deepseek-R1 1.5B & 0.000 & 0.000 & 0.000 & 0.000 & 0.000 \\
Llama3.2 1B & 0.009 & 0.018 & 0.018 & 0.019 & 0.000 \\
\bottomrule
\end{tabular}
\end{table}

\begin{table}[!htbp]
\caption{Performance of French Full Manual Models}
\label{tab:french_full_manual_performance}
  \tiny
\begin{tabular}{lccccc}
\toprule
Metric & Acc. & F1 & Prec. & Rec. & Spec. \\
LLM &  &  &  &  &  \\
\midrule
\midrule
Gemini 2.5 Flash & \textbf{0.704} & \textbf{0.754} & 0.645 & \textbf{0.907} & 0.500 \\
Qwen 2.5 7B & 0.620 & 0.549 & \textbf{0.676} & 0.463 & \textbf{0.778} \\
Qwen3 8B & 0.398 & 0.414 & 0.404 & 0.426 & 0.370 \\
Phi3 14B & 0.537 & 0.528 & 0.538 & 0.519 & 0.556 \\
Llama3.1 8B & 0.148 & 0.193 & 0.183 & 0.204 & 0.093 \\
Deepseek-R1 8B & 0.259 & 0.310 & 0.290 & 0.333 & 0.185 \\
Llama3.2 3B & 0.019 & 0.036 & 0.036 & 0.037 & 0.000 \\
Deepseek-R1 1.5B & 0.009 & 0.018 & 0.018 & 0.019 & 0.000 \\
Llama3.2 1B & 0.000 & 0.000 & 0.000 & 0.000 & 0.000 \\
\bottomrule
\end{tabular}
\end{table}

\begin{table}[!htbp]
\caption{Performance of German Full Manual Models}
\label{tab:german_full_manual_performance}
  \tiny
\begin{tabular}{lccccc}
\toprule
Metric & Acc. & F1 & Prec. & Rec. & Spec. \\
LLM &  &  &  &  &  \\
\midrule
\midrule
Gemini 2.5 Flash & \textbf{0.685} & \textbf{0.738} & \textbf{0.632} & \textbf{0.889} & 0.481 \\
Qwen 2.5 7B & 0.546 & 0.380 & 0.600 & 0.278 & \textbf{0.815} \\
Qwen3 8B & 0.352 & 0.364 & 0.357 & 0.370 & 0.333 \\
Phi3 14B & 0.519 & 0.480 & 0.522 & 0.444 & 0.593 \\
Llama3.1 8B & 0.194 & 0.269 & 0.246 & 0.296 & 0.093 \\
Deepseek-R1 8B & 0.231 & 0.303 & 0.277 & 0.333 & 0.130 \\
Llama3.2 3B & 0.037 & 0.071 & 0.069 & 0.074 & 0.000 \\
Deepseek-R1 1.5B & 0.056 & 0.105 & 0.100 & 0.111 & 0.000 \\
Llama3.2 1B & 0.019 & 0.036 & 0.036 & 0.037 & 0.000 \\
\bottomrule
\end{tabular}
\end{table}

{
\FloatBarrier 
\raggedbottom 



\subsection*{Performance Heatmaps: Multilingual Algorithm vs. Model}

\begin{figure}[H] 
    \centering
    \includegraphics[width=\columnwidth]{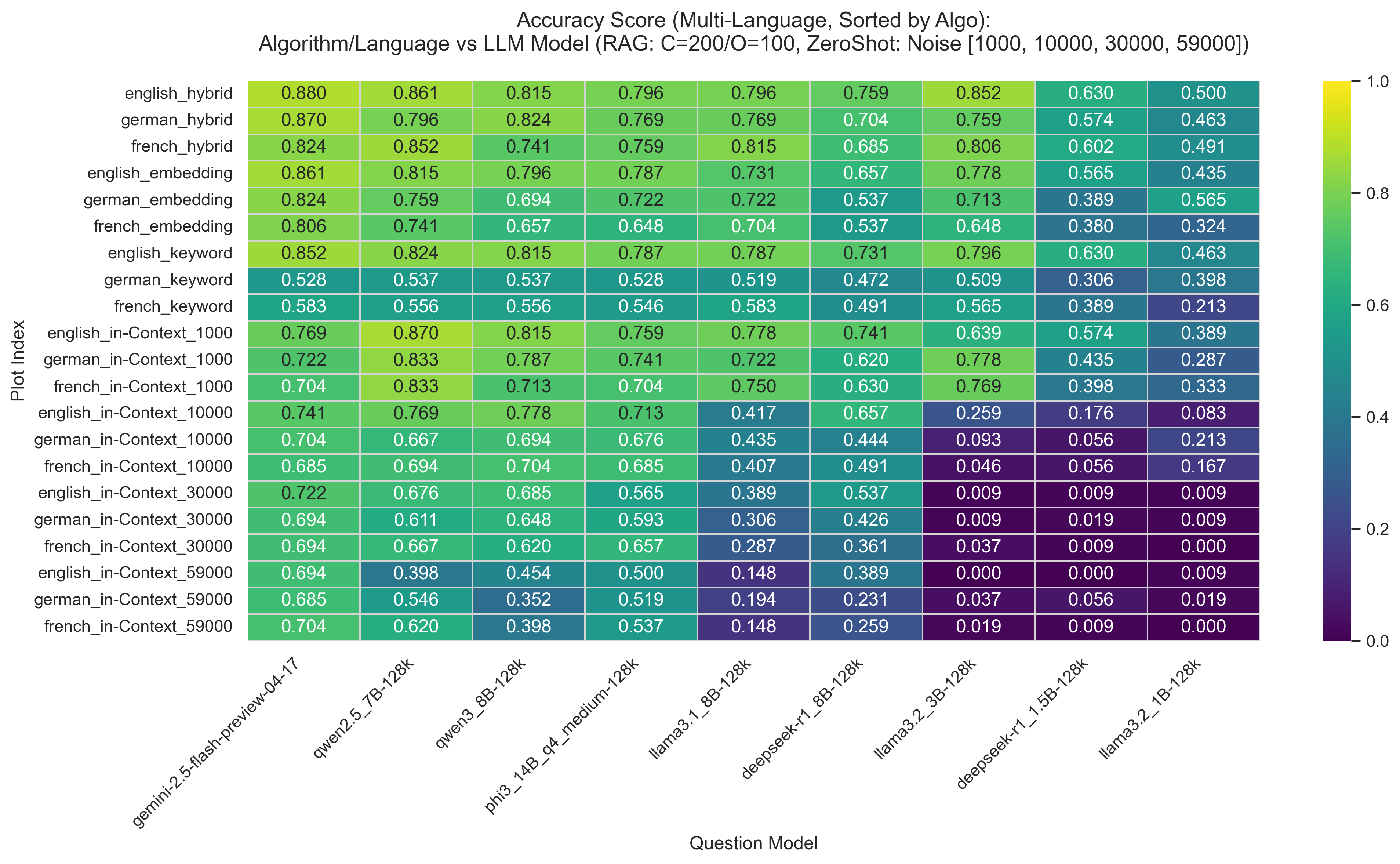}
    \caption{Accuracy score heatmap comparing multilingual algorithms and models.}
    \label{fig:acc_heatmap_multilang_algo_model}
\end{figure}

\begin{figure}[H] 
    \centering
    \includegraphics[width=\columnwidth]{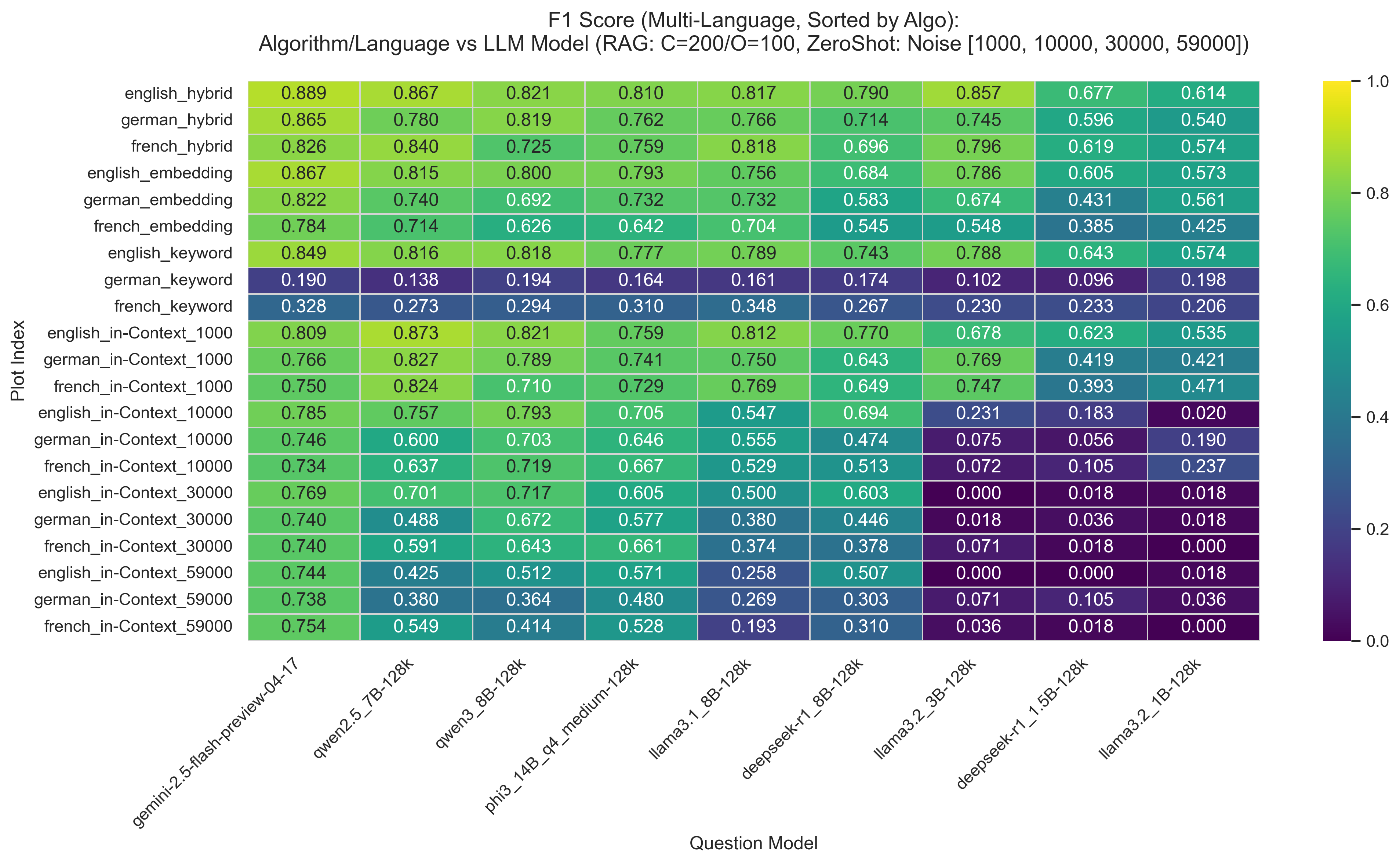}
    \caption{F1 score heatmap comparing multilingual algorithms and models.}
    \label{fig:f1_heatmap_multilang_algo_model}
\end{figure}

\begin{figure}[H] 
    \centering
    \includegraphics[width=\columnwidth]{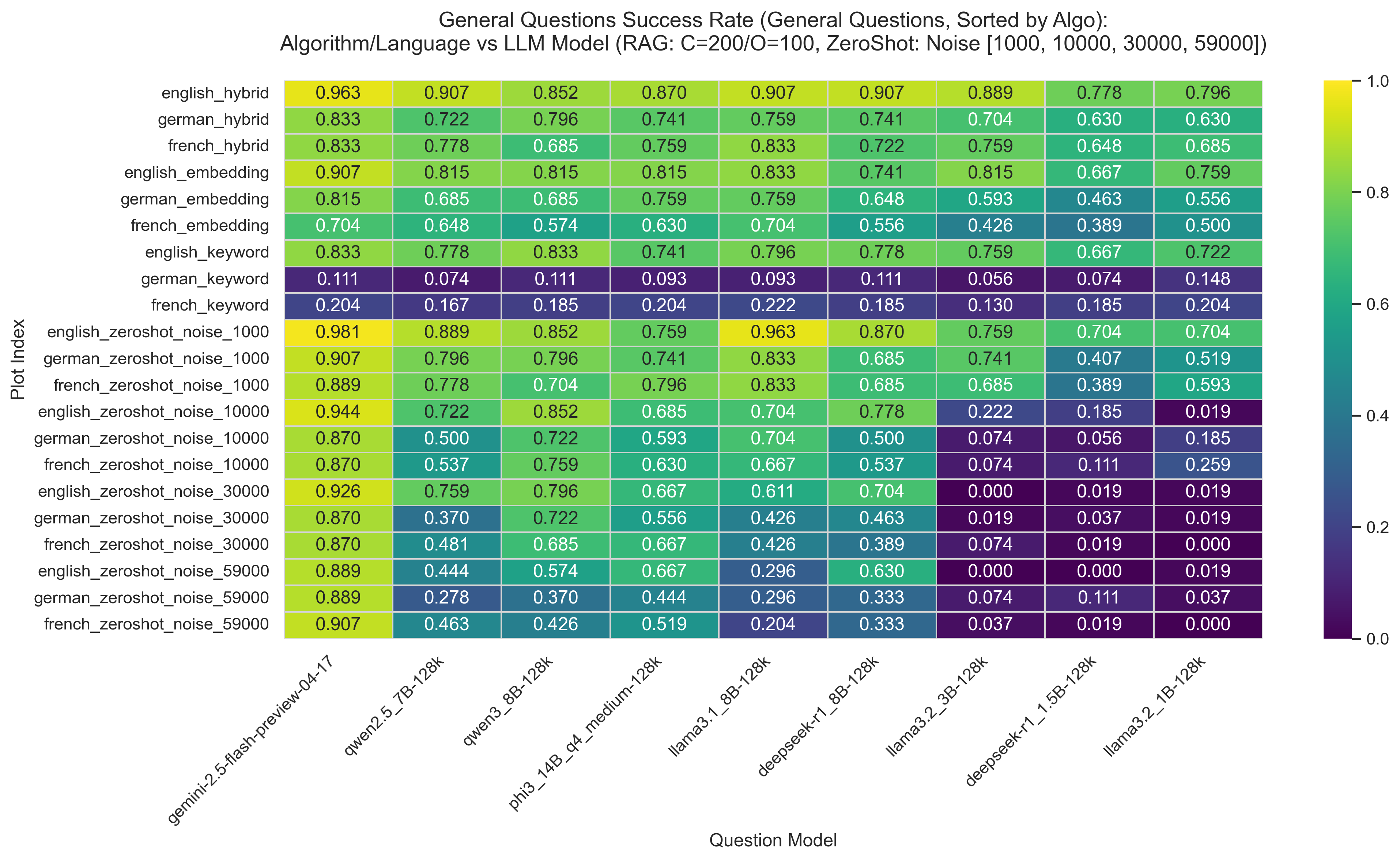}
    \caption{Success heatmap for general questions: multilingual algorithms vs. models.}
    \label{fig:gen_q_heatmap_multilang_algo_model}
\end{figure}

\begin{figure}[H]
    \centering
    \includegraphics[width=\columnwidth]{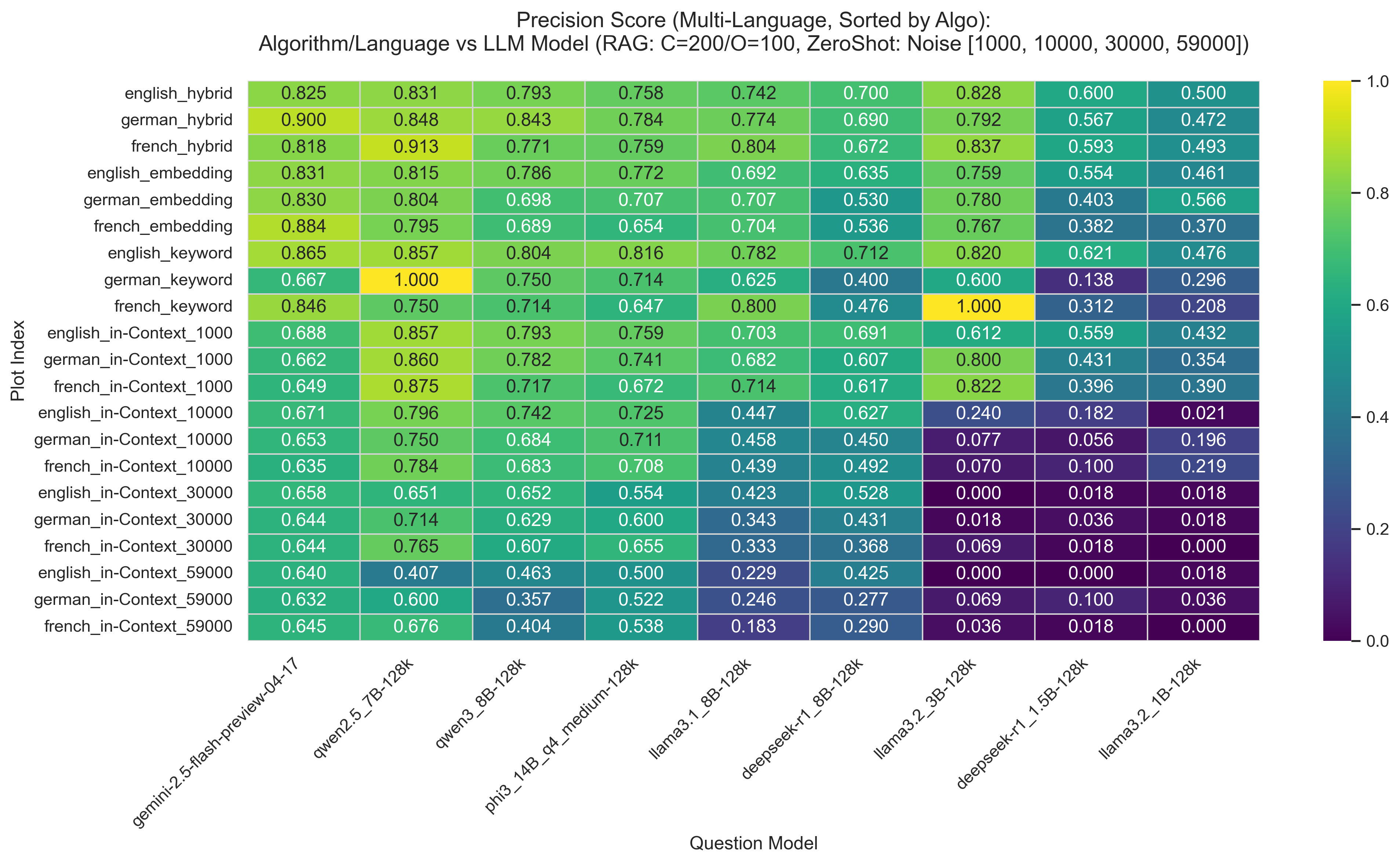}
    \caption{Precision score heatmap comparing multilingual algorithms and models.}
    \label{fig:prec_heatmap_multilang_algo_model}
\end{figure}

\begin{figure}[H]
    \centering
    \includegraphics[width=\columnwidth]{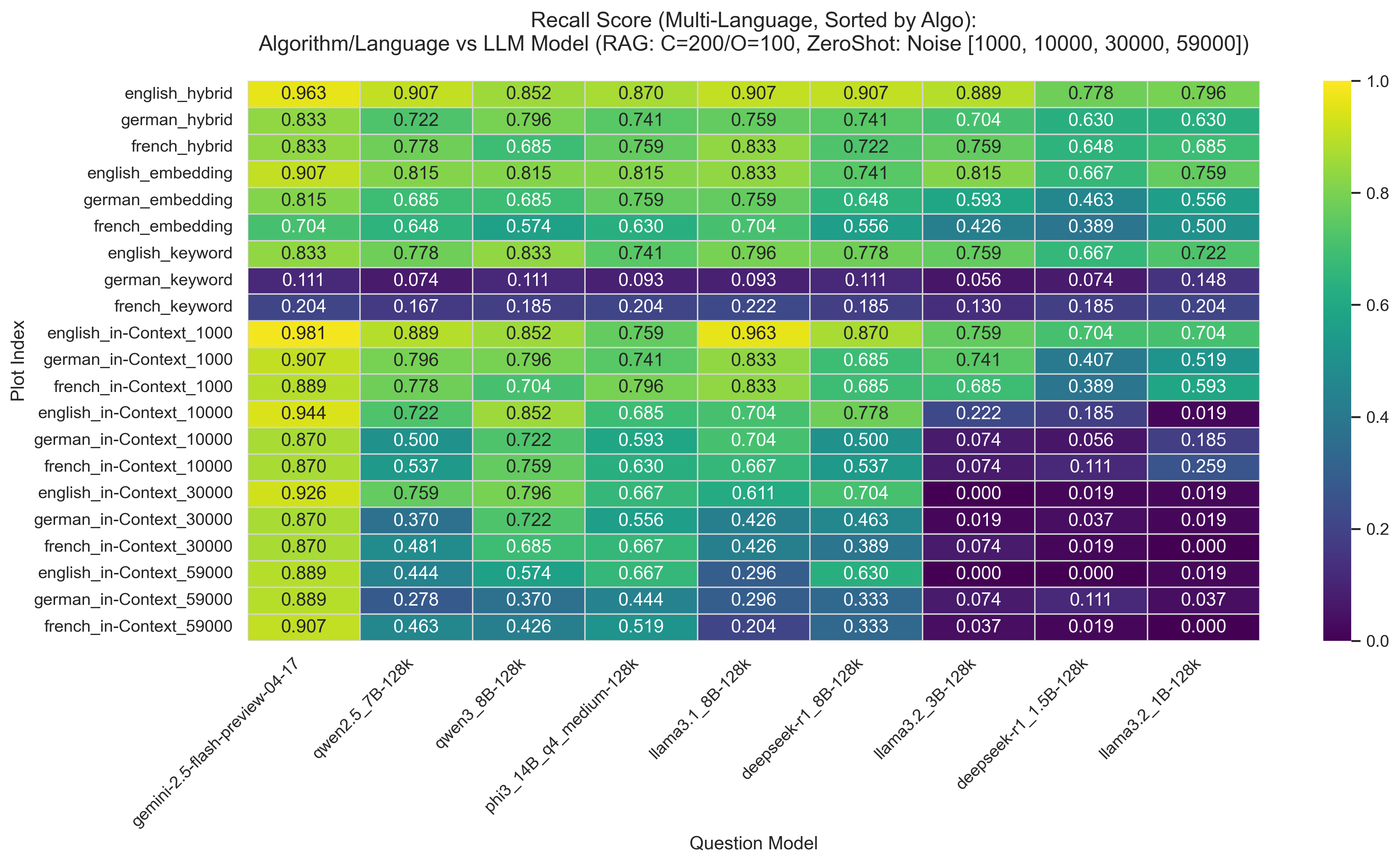}
    \caption{Recall score heatmap comparing multilingual algorithms and models.}
    \label{fig:recall_heatmap_multilang_algo_model}
\end{figure}

\begin{figure}[H]
    \centering
    \includegraphics[width=\columnwidth]{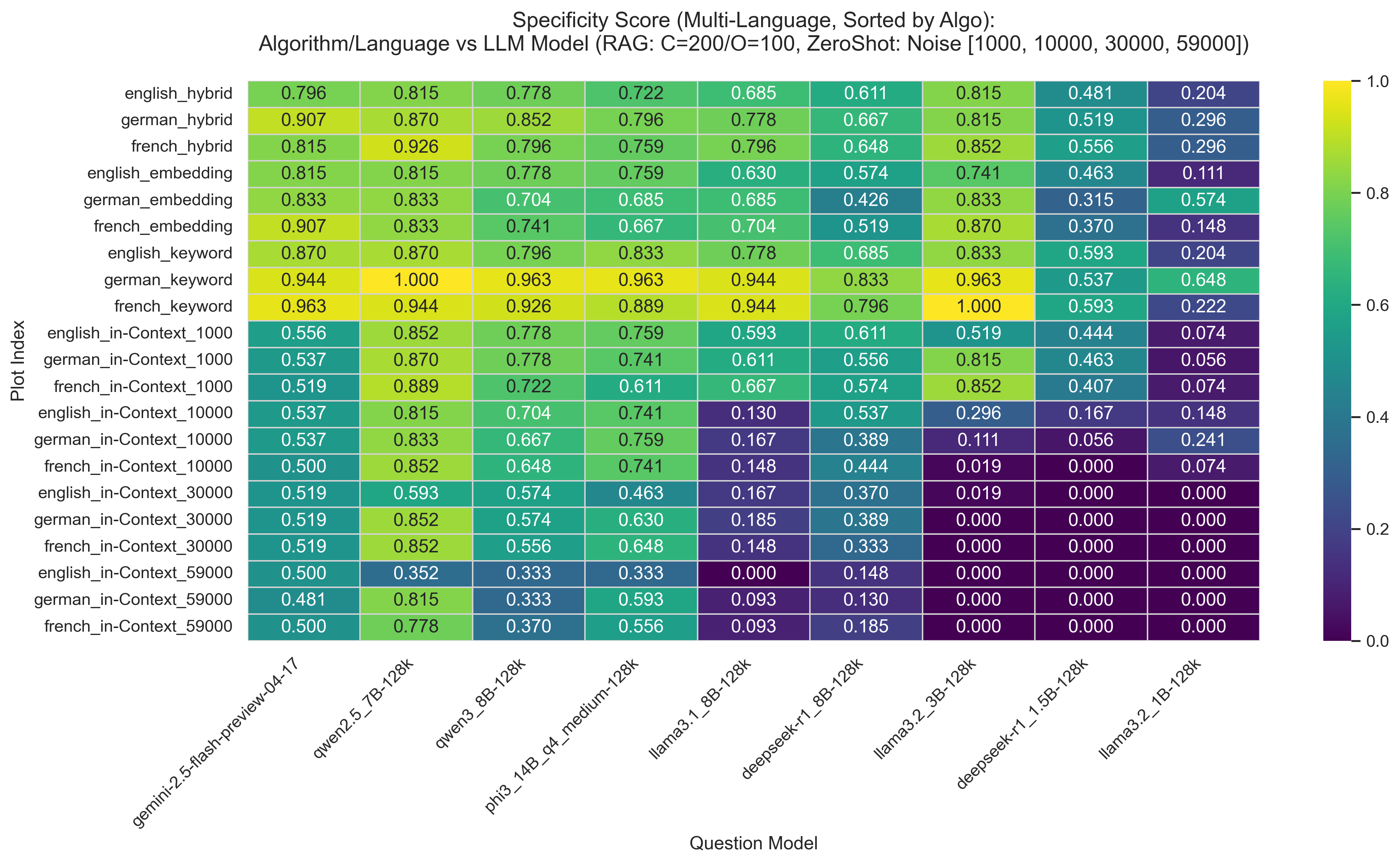}
    \caption{Specificity score heatmap comparing multilingual algorithms and models.}
    \label{fig:spec_heatmap_multilang_algo_model}
\end{figure}

\begin{figure}[H]
    \centering
    \includegraphics[width=\columnwidth]{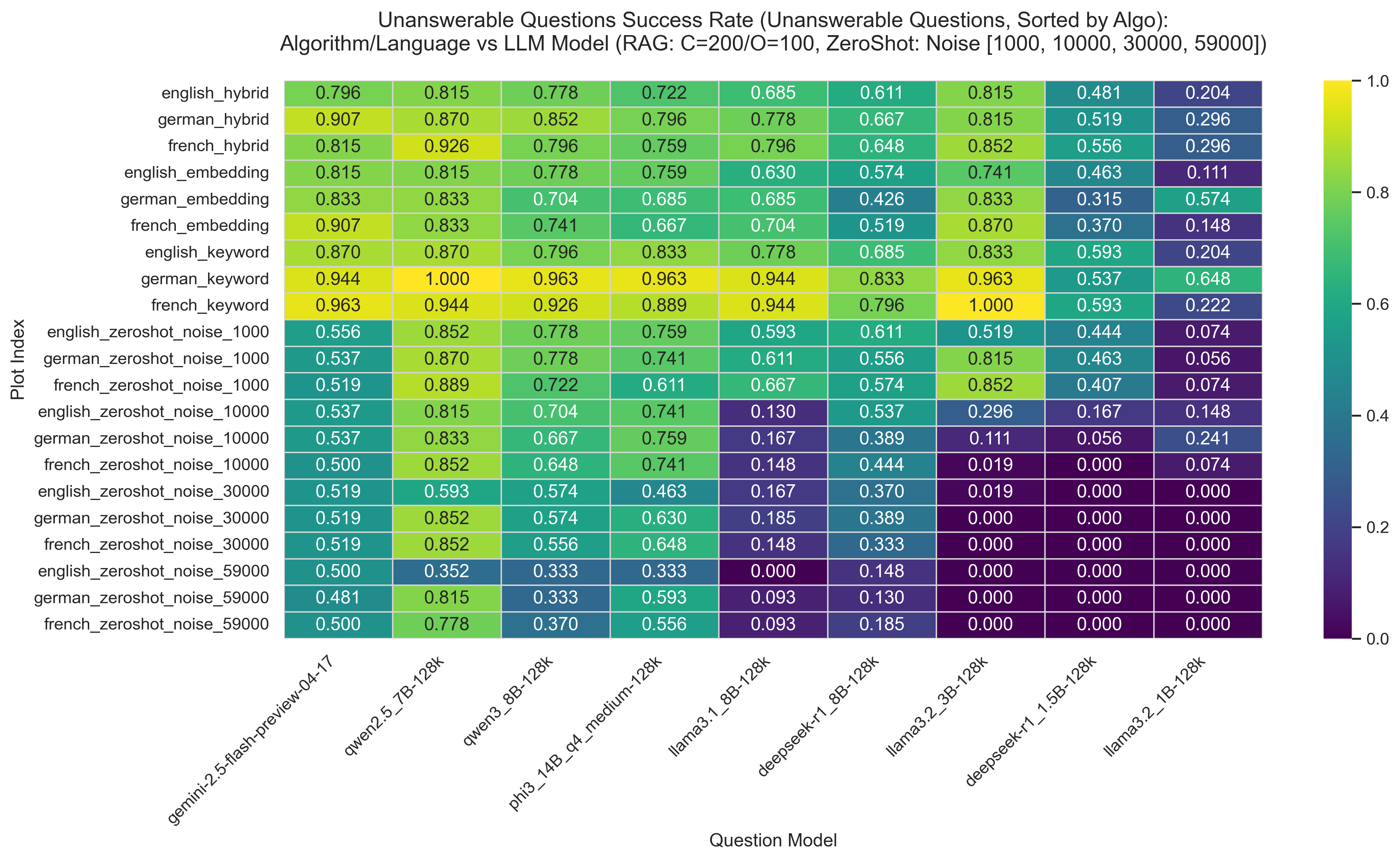}
    \caption{Success heatmap for unanswerable questions: multilingual algorithms vs. models.}
    \label{fig:unans_q_heatmap_multilang_algo_model}
\end{figure}


\subsection*{Long-Context QA Performance vs. Noise}

\begin{figure}[H]
    \centering
    \includegraphics[width=\columnwidth]{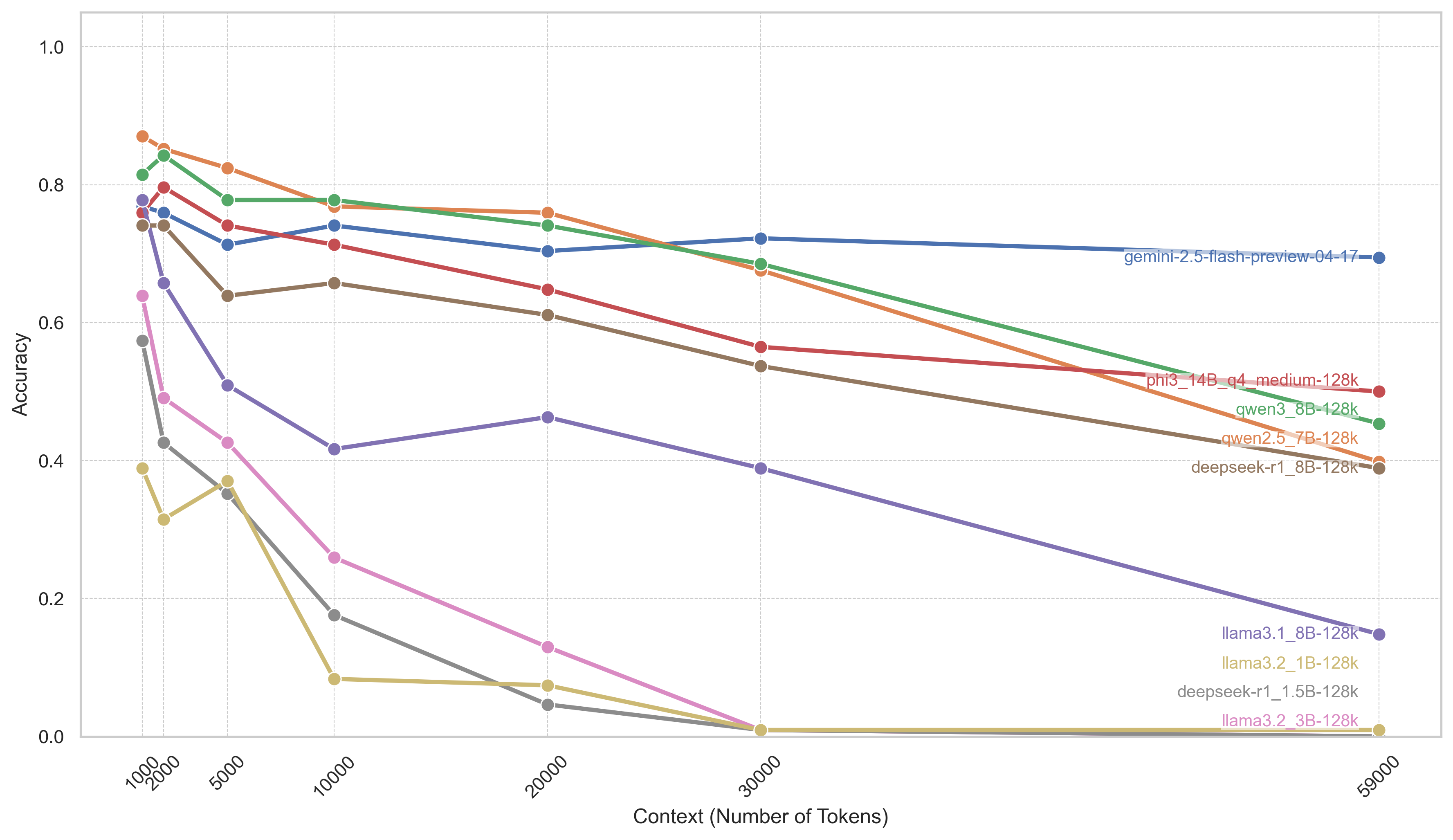}
    \caption{Long-Context QA accuracy vs. noise: English.}
    \label{fig:zs_acc_noise_en}
\end{figure}

\begin{figure}[H]
    \centering
    \includegraphics[width=\columnwidth]{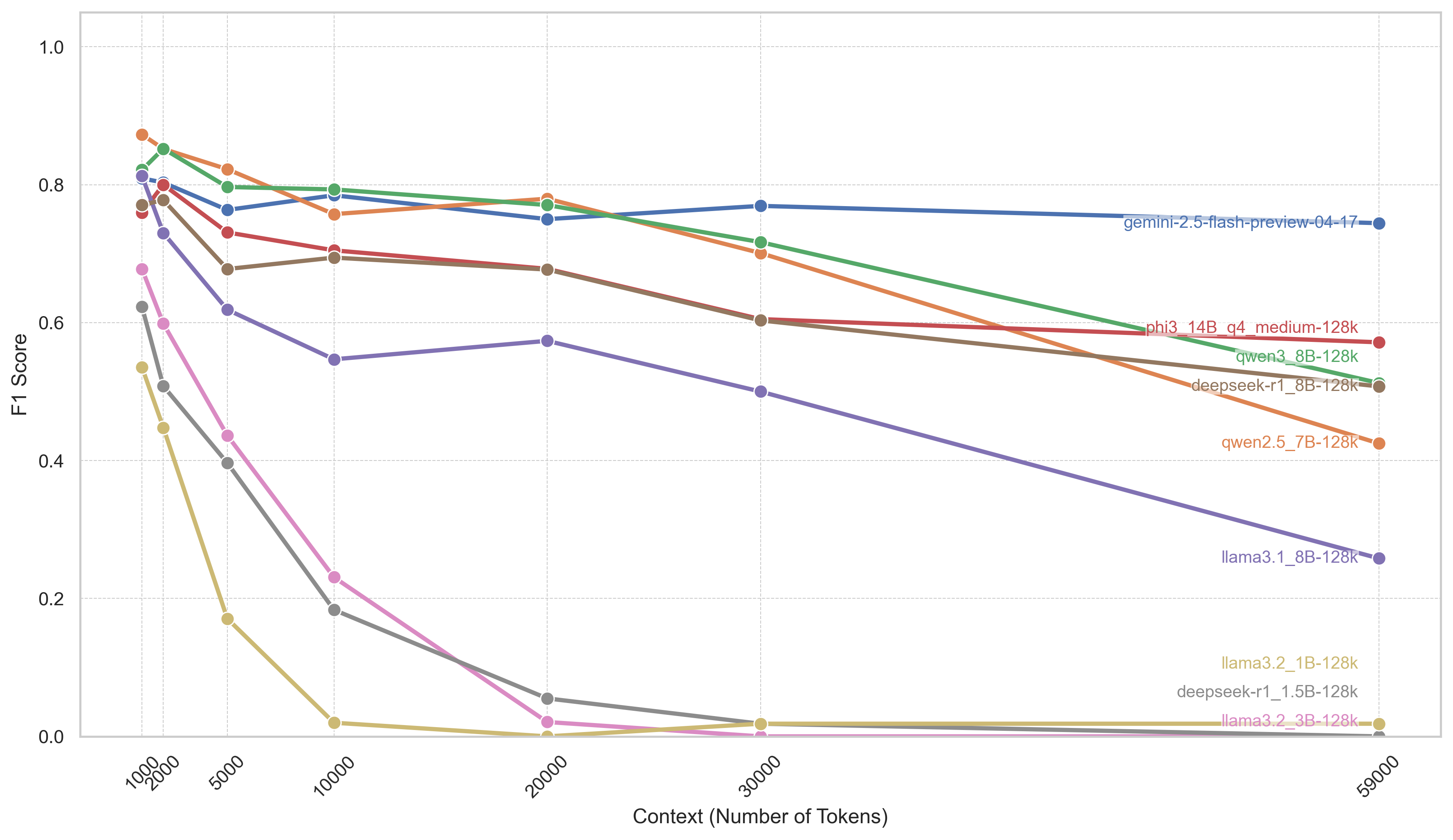}
    \caption{Long-Context QA F1 score vs. noise: English.}
    \label{fig:zs_f1_noise_en}
\end{figure}

\begin{figure}[H]
    \centering
    \includegraphics[width=\columnwidth]{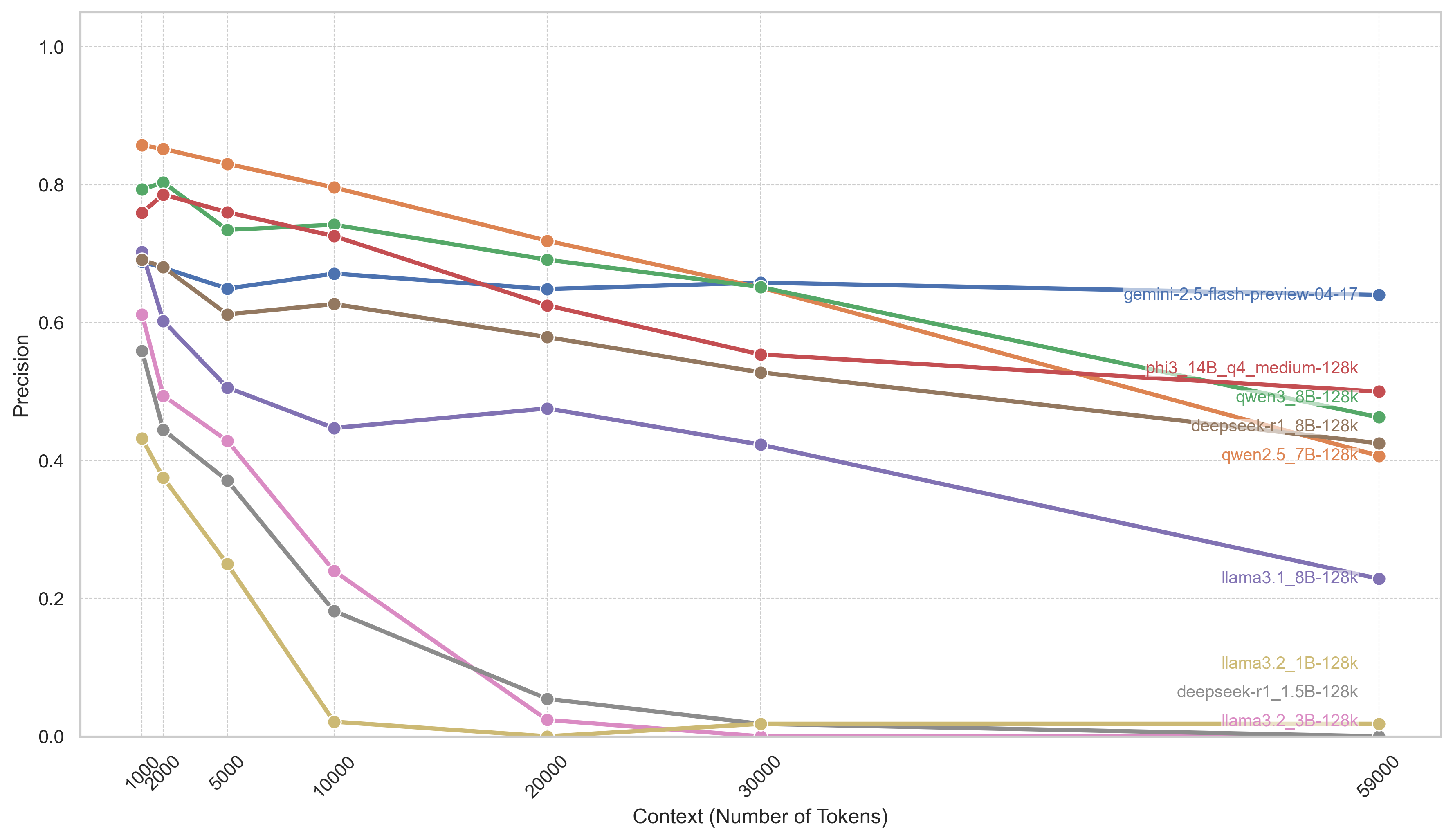}
    \caption{Long-Context QA precision vs. noise: English.}
    \label{fig:zs_prec_noise_en}
\end{figure}

\begin{figure}[H]
    \centering
    \includegraphics[width=\columnwidth]{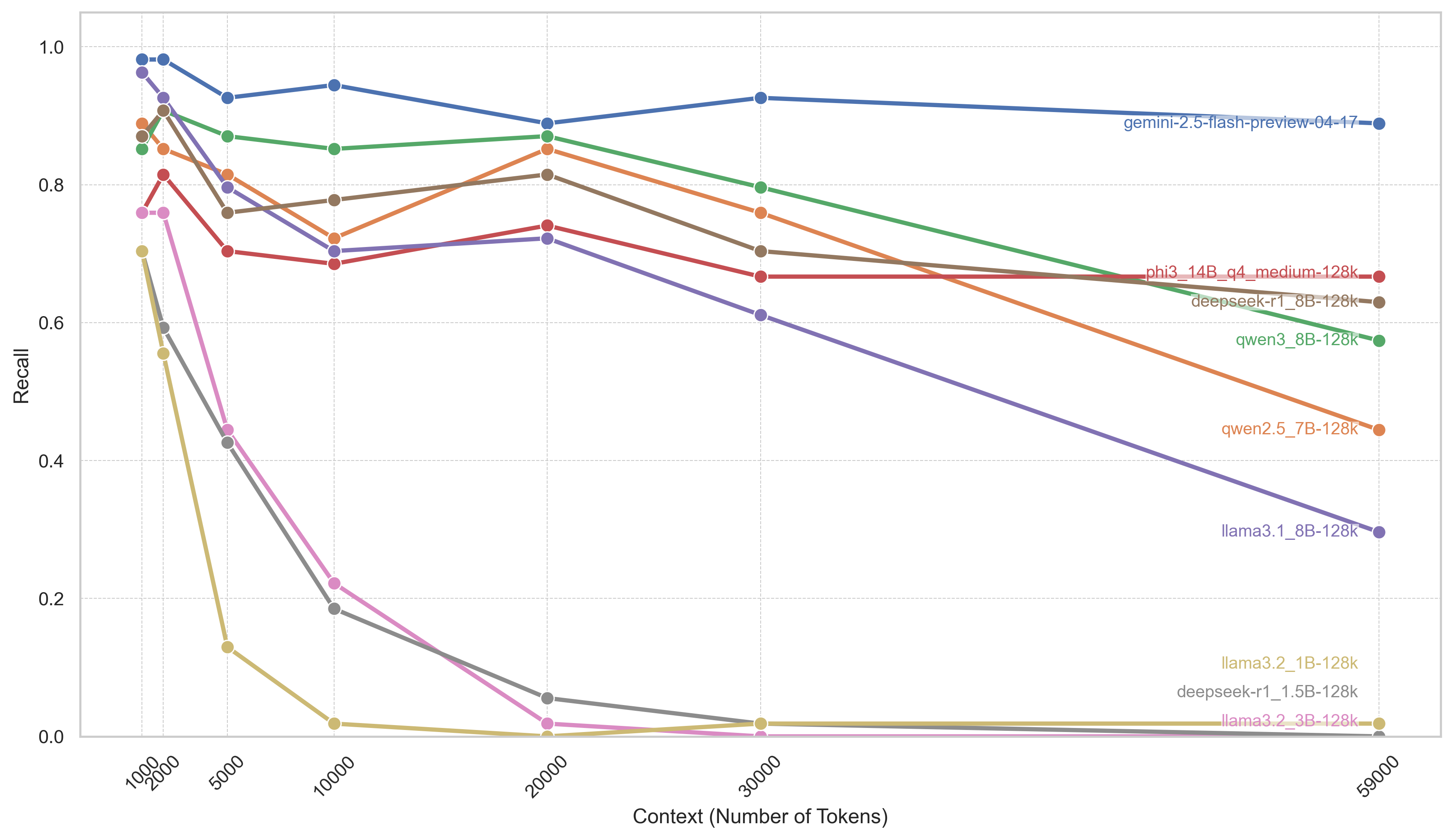}
    \caption{Long-Context QA recall vs. noise: English.}
    \label{fig:zs_recall_noise_en}
\end{figure}

\begin{figure}[H]
    \centering
    \includegraphics[width=\columnwidth]{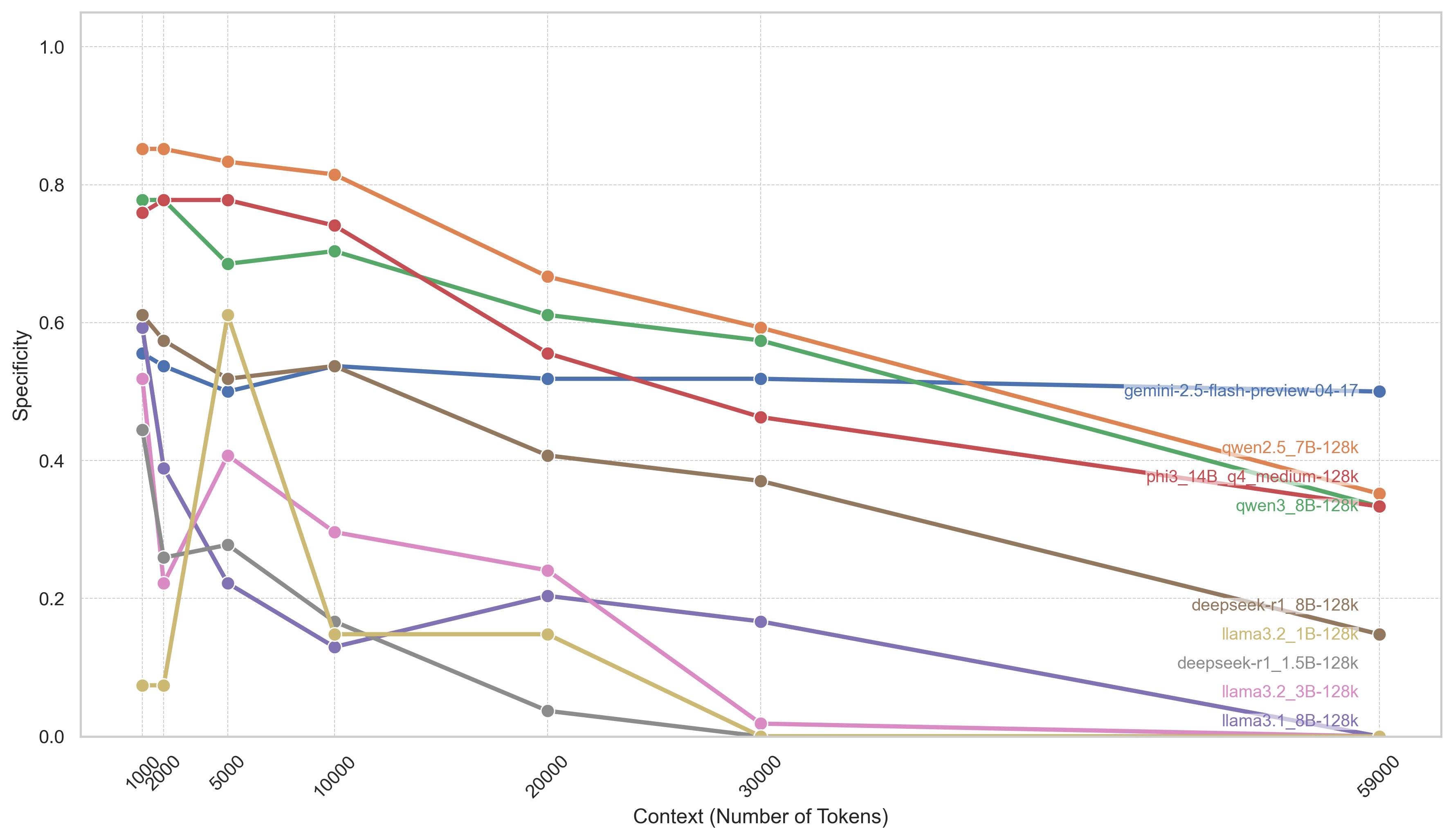}
    \caption{Long-Context QA specificity vs. noise: English.}
    \label{fig:zs_spec_noise_en}
\end{figure}


\begin{figure}[H]
    \centering
    \includegraphics[width=\columnwidth]{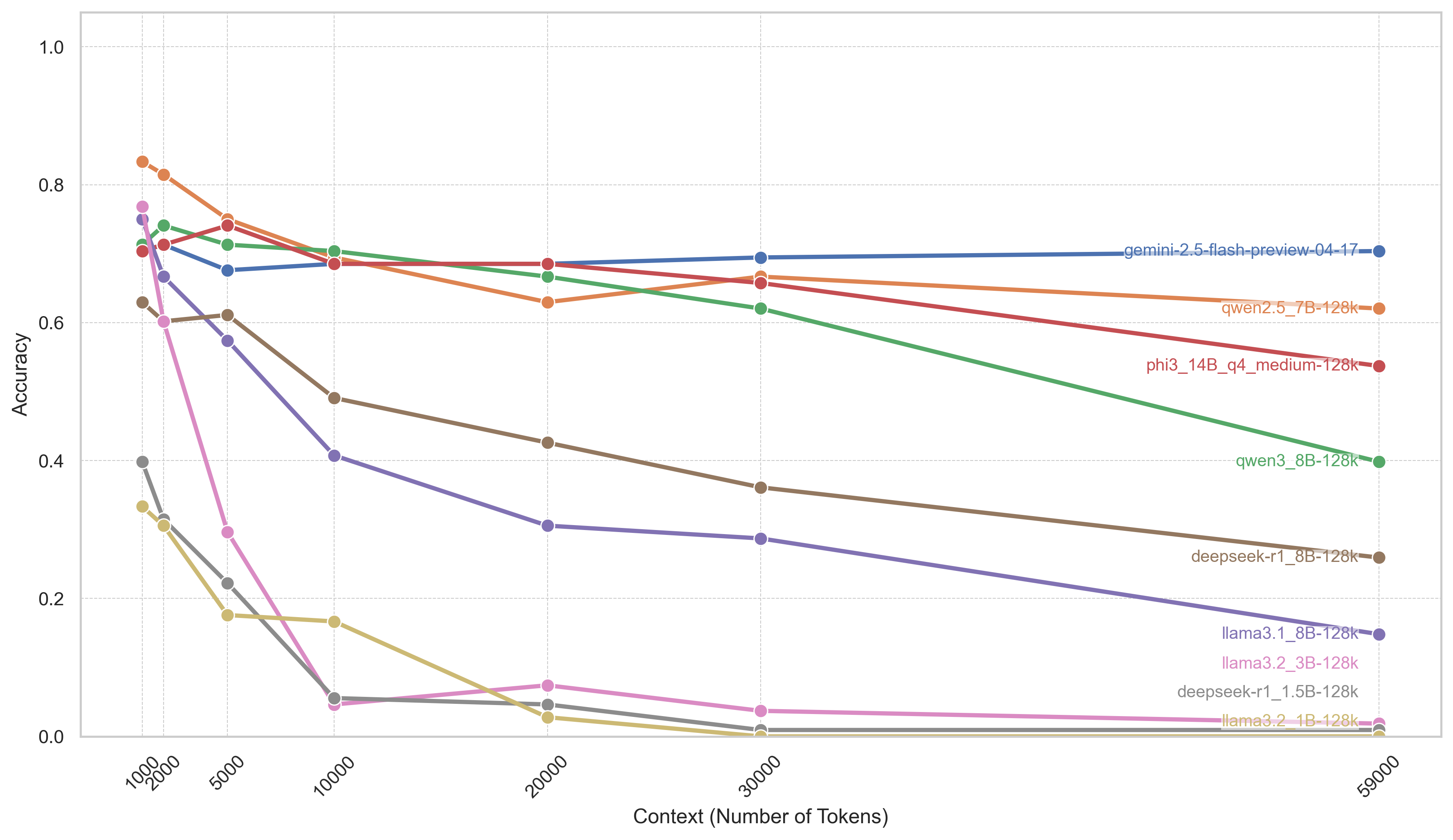}
    \caption{Long-Context QA accuracy vs. noise: French.}
    \label{fig:zs_acc_noise_fr}
\end{figure}

\begin{figure}[H]
    \centering
    \includegraphics[width=\columnwidth]{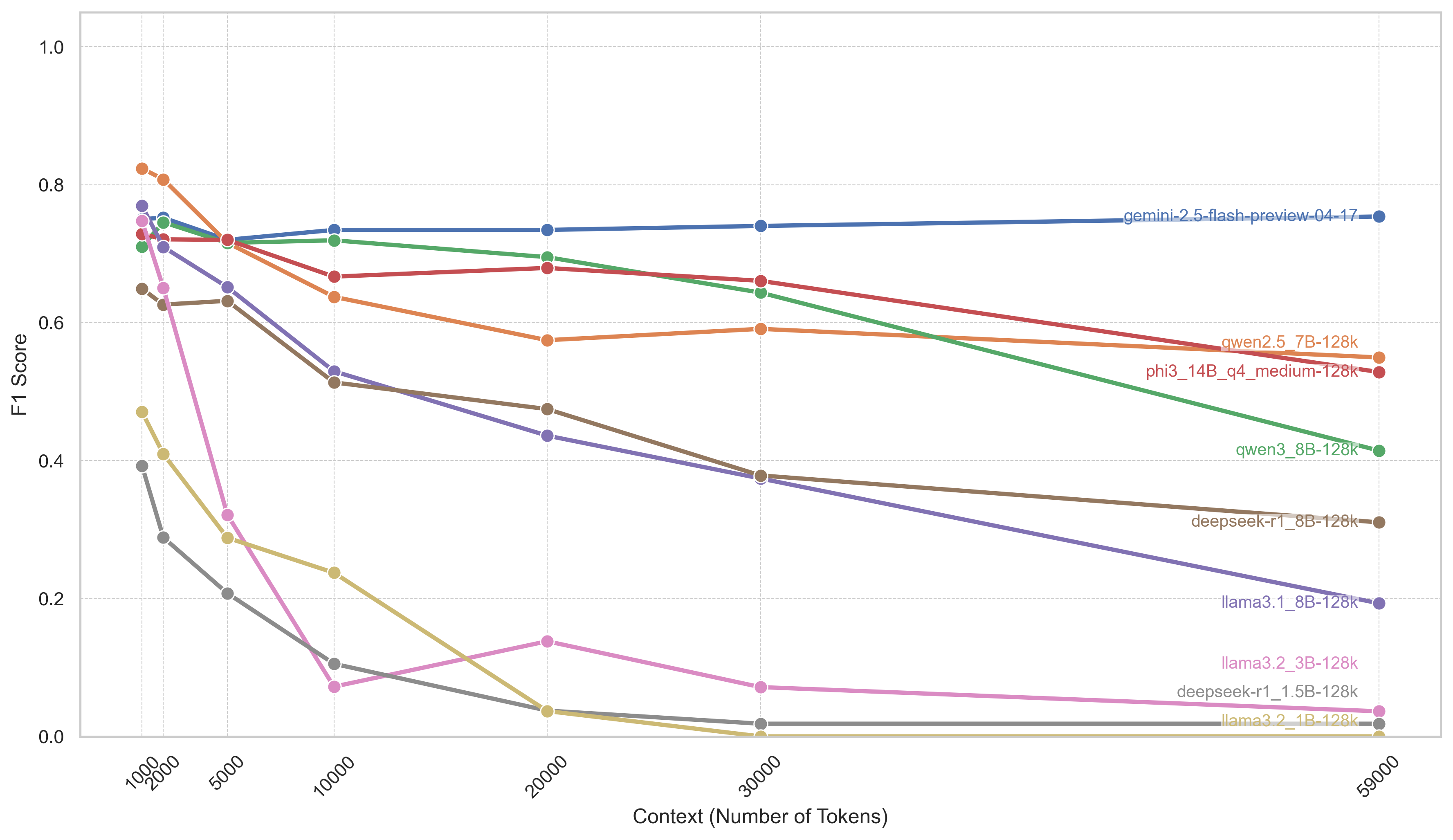}
    \caption{Long-Context QA F1 score vs. noise: French.}
    \label{fig:zs_f1_noise_fr}
\end{figure}

\begin{figure}[H]
    \centering
    \includegraphics[width=\columnwidth]{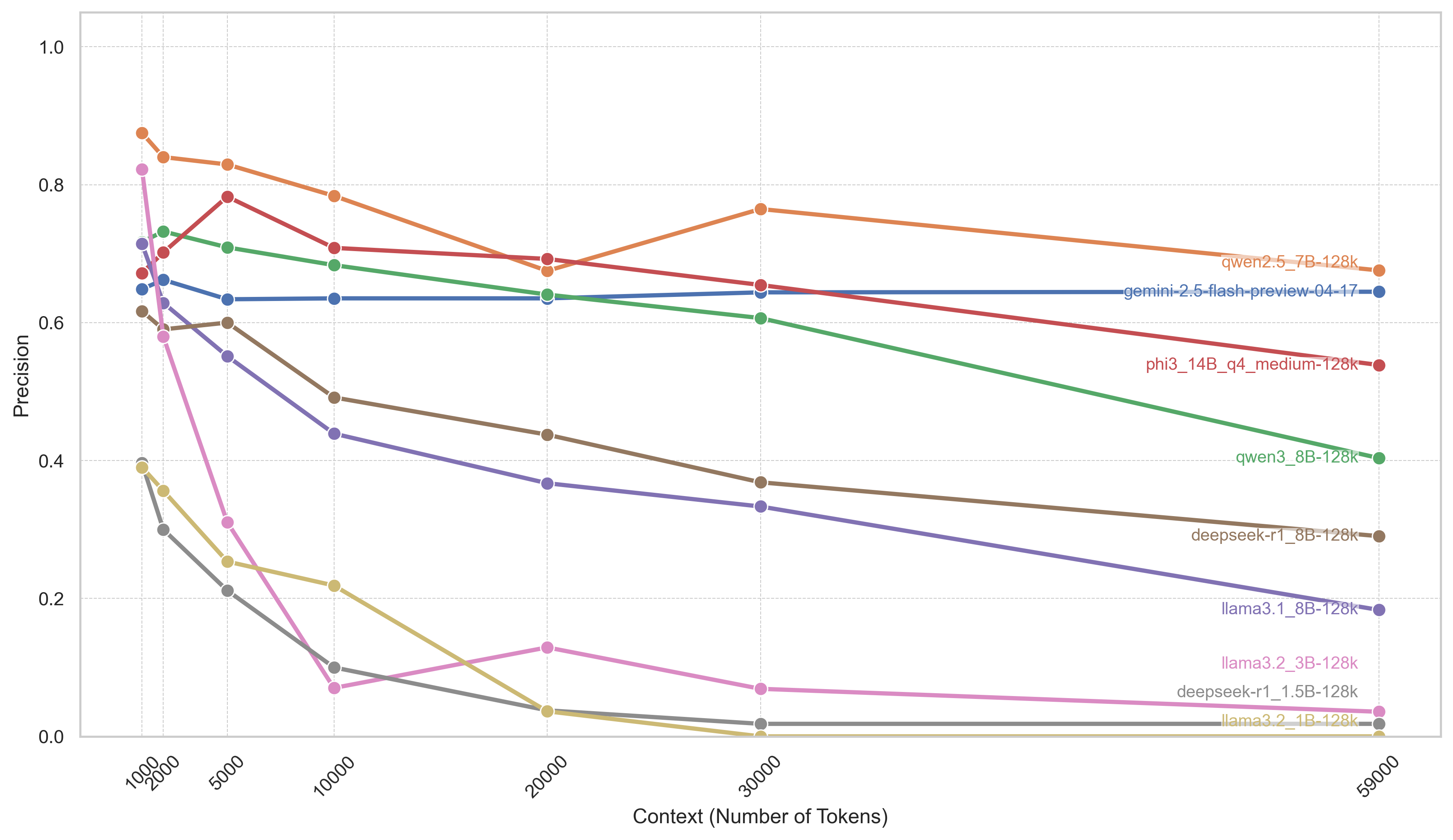}
    \caption{Long-Context QA precision vs. noise: French.}
    \label{fig:zs_prec_noise_fr}
\end{figure}

\begin{figure}[H]
    \centering
    \includegraphics[width=\columnwidth]{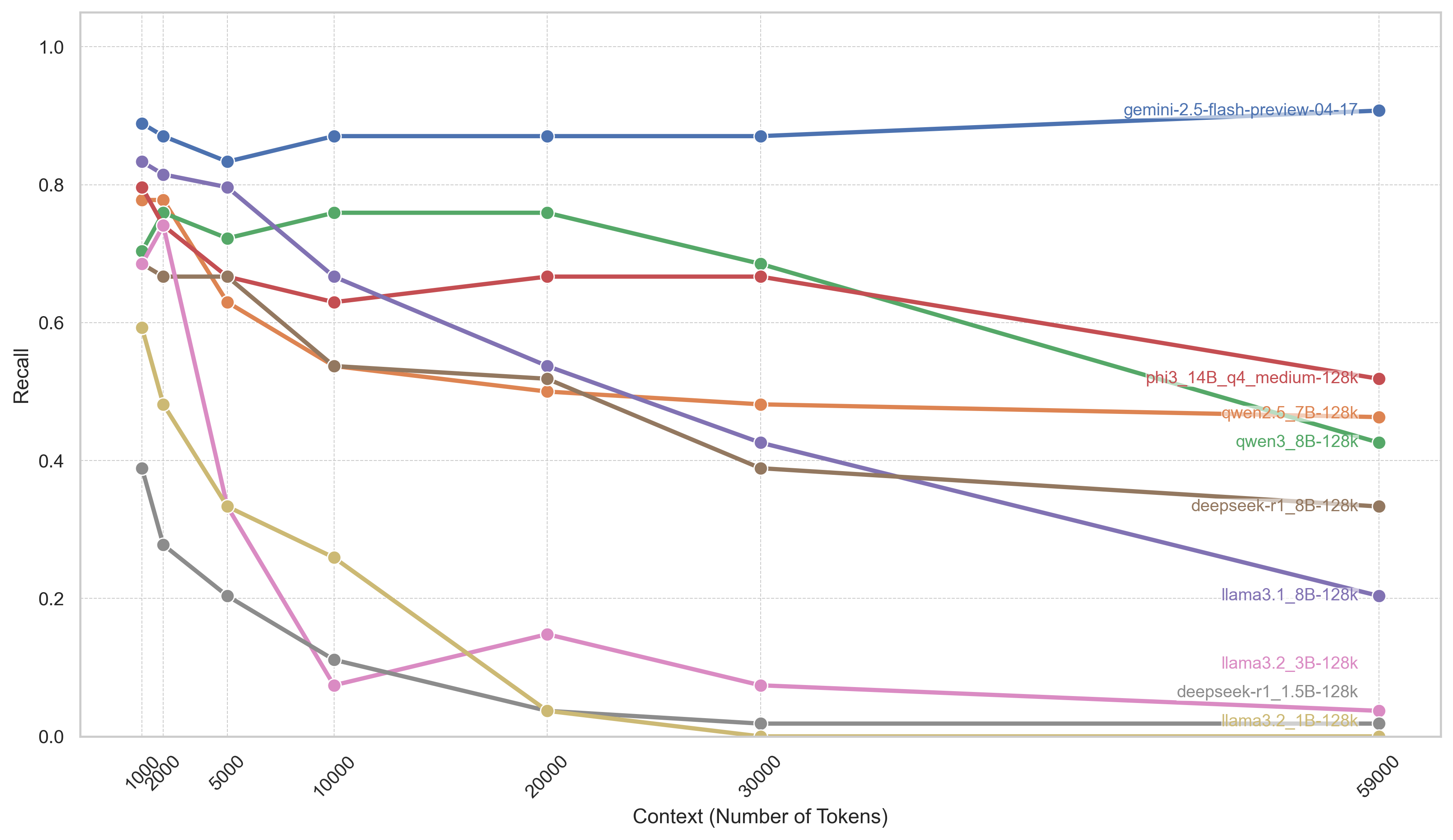}
    \caption{Long-Context QA recall vs. noise: French.}
    \label{fig:zs_recall_noise_fr}
\end{figure}

\begin{figure}[H]
    \centering
    \includegraphics[width=\columnwidth]{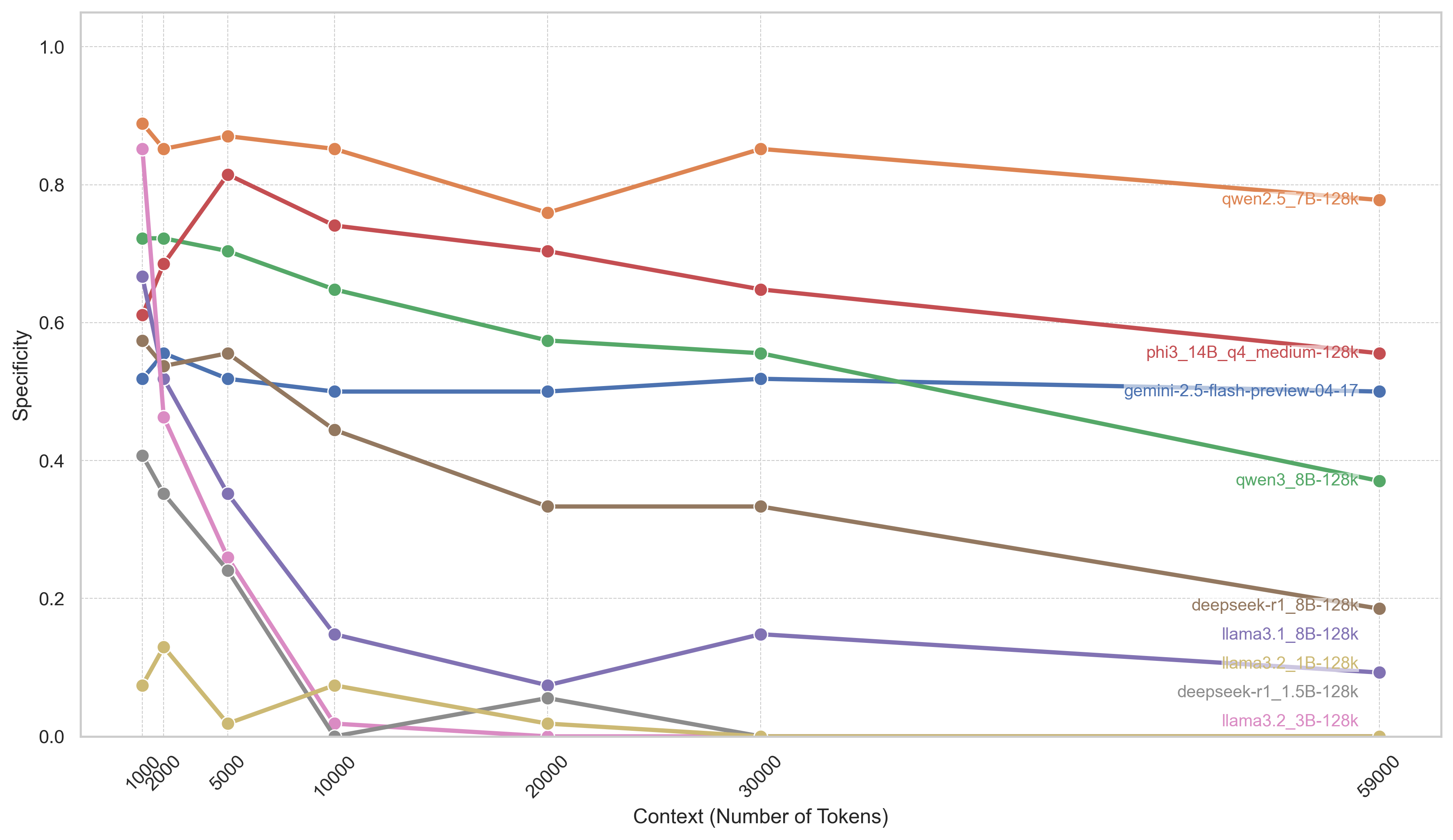}
    \caption{Long-Context QA specificity vs. noise: French.}
    \label{fig:zs_spec_noise_fr}
\end{figure}


\begin{figure}[H]
    \centering
    \includegraphics[width=\columnwidth]{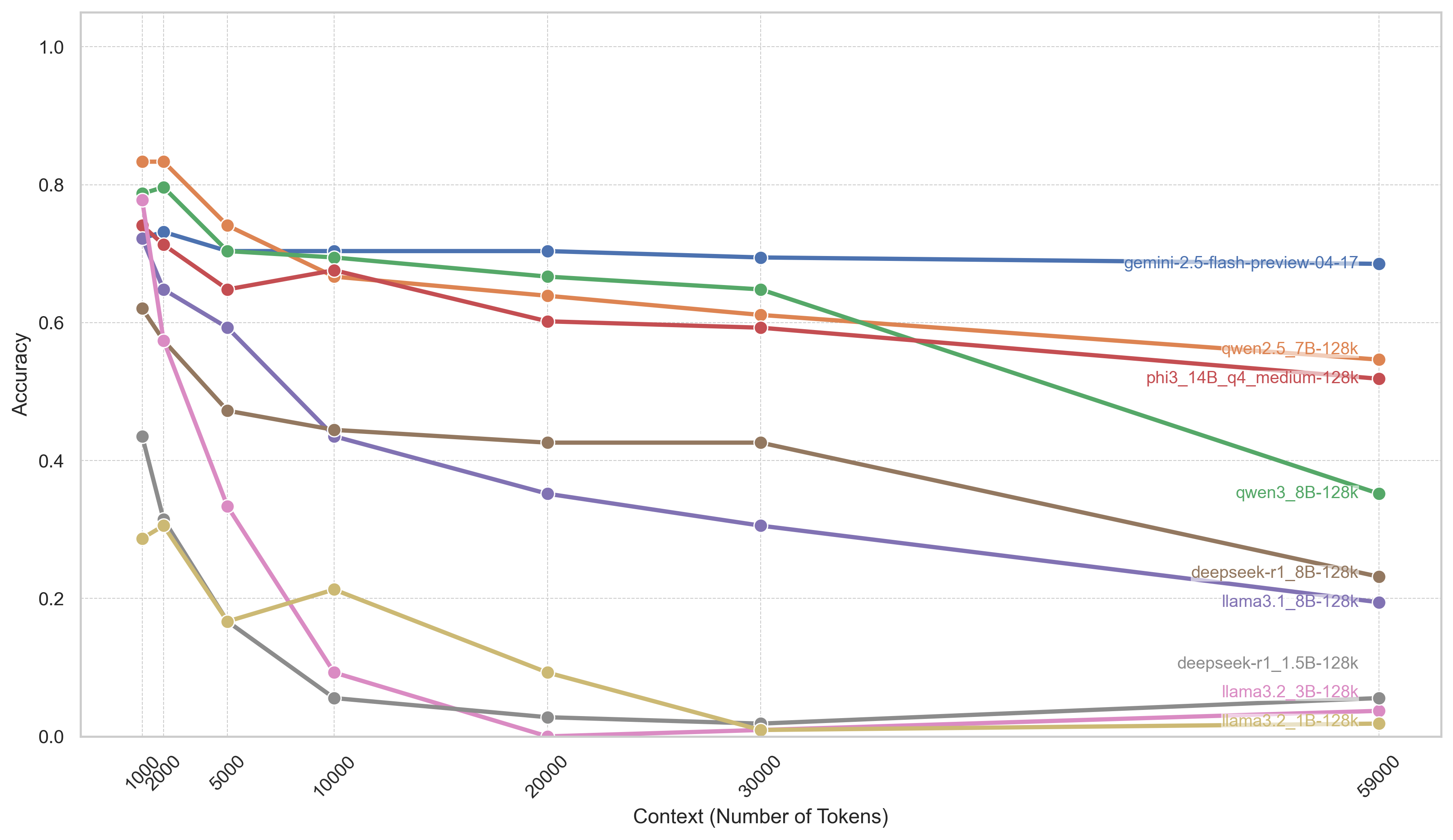}
    \caption{Long-Context QA accuracy vs. noise: German.}
    \label{fig:zs_acc_noise_de}
\end{figure}

\begin{figure}[H]
    \centering
    \includegraphics[width=\columnwidth]{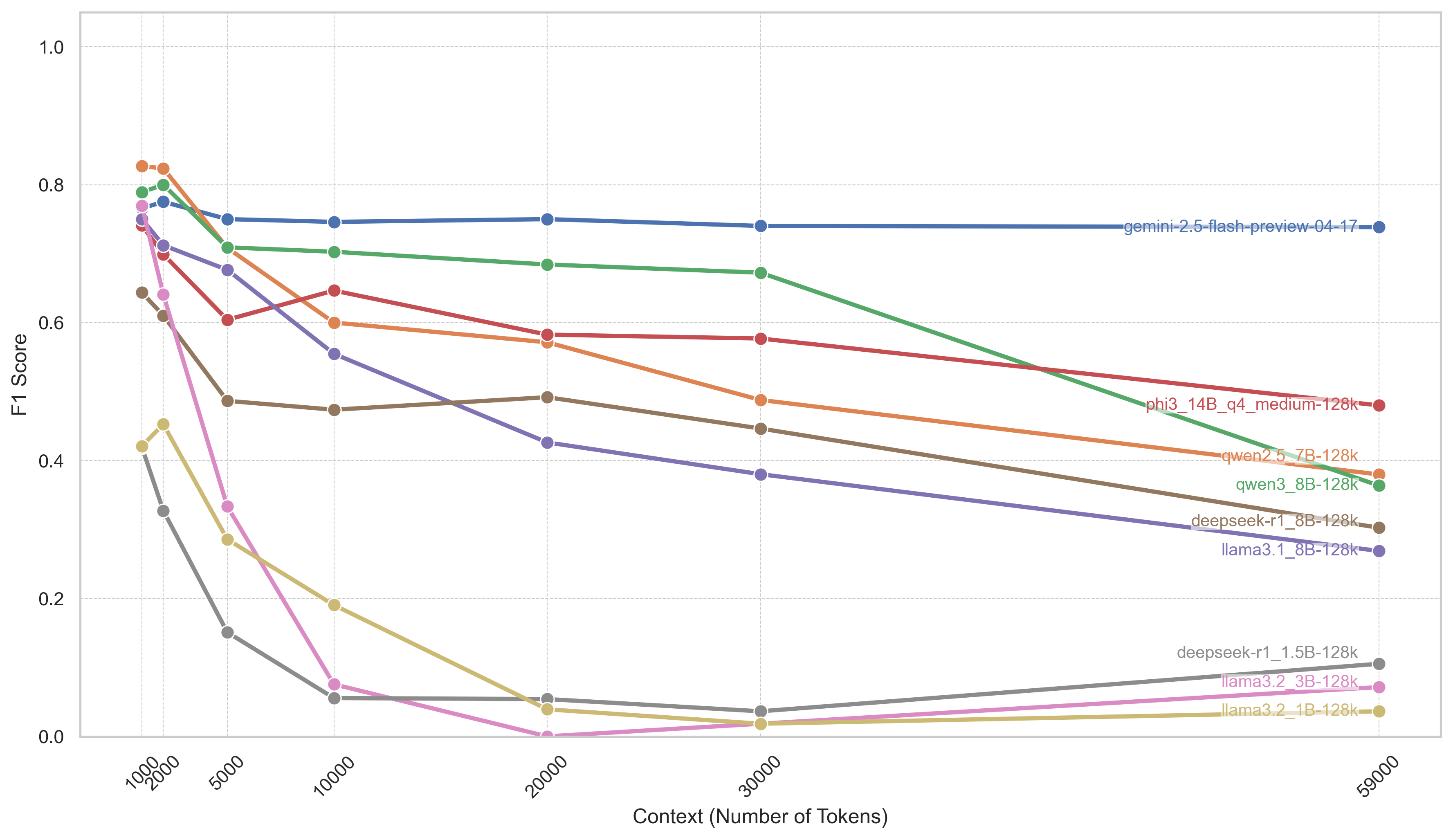}
    \caption{Long-Context QA F1 score vs. noise: German.}
    \label{fig:zs_f1_noise_de}
\end{figure}

\begin{figure}[H]
    \centering
    \includegraphics[width=\columnwidth]{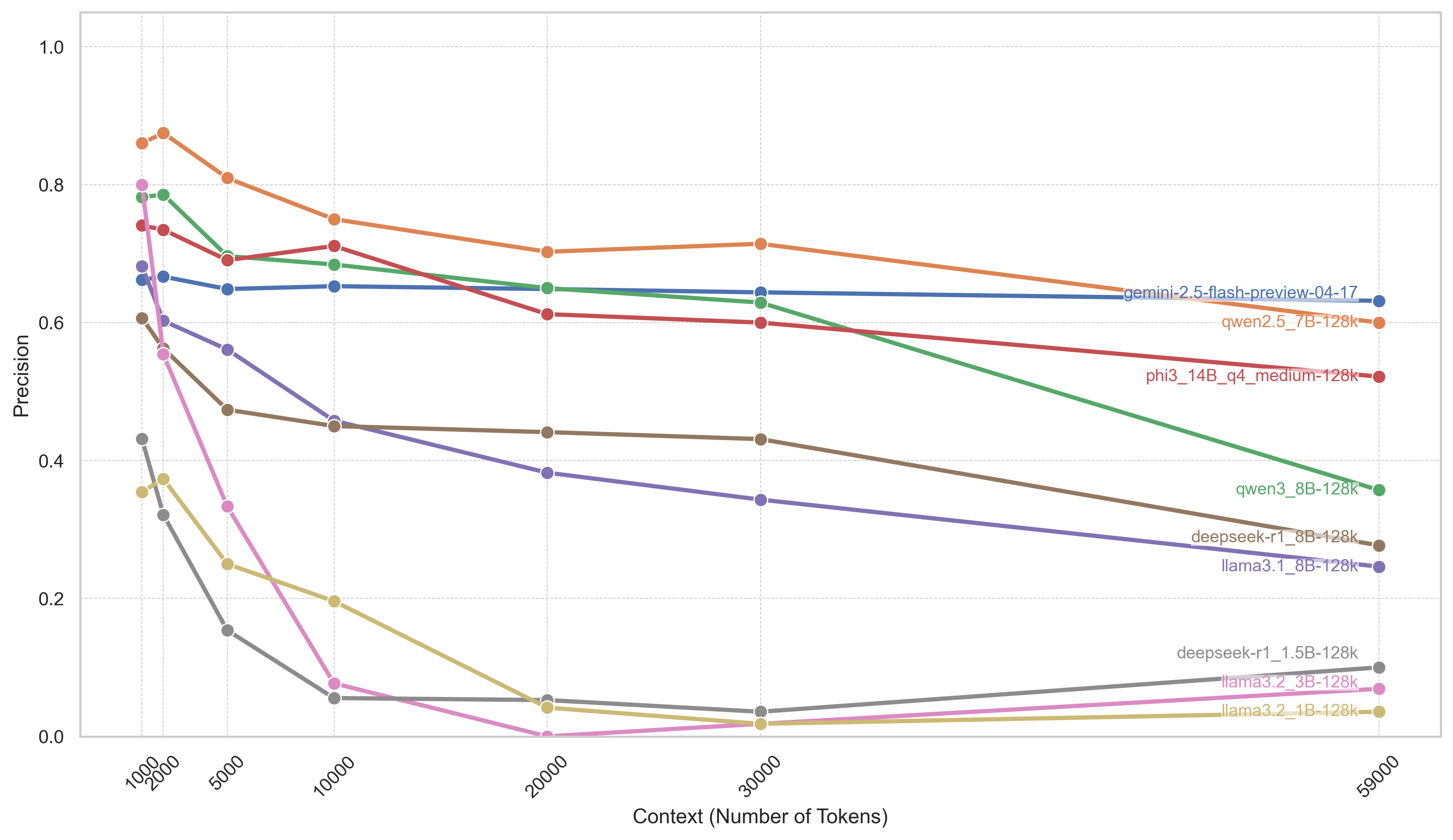}
    \caption{Long-Context QA precision vs. noise: German.}
    \label{fig:zs_prec_noise_de}
\end{figure}

\begin{figure}[H]
    \centering
    \includegraphics[width=\columnwidth]{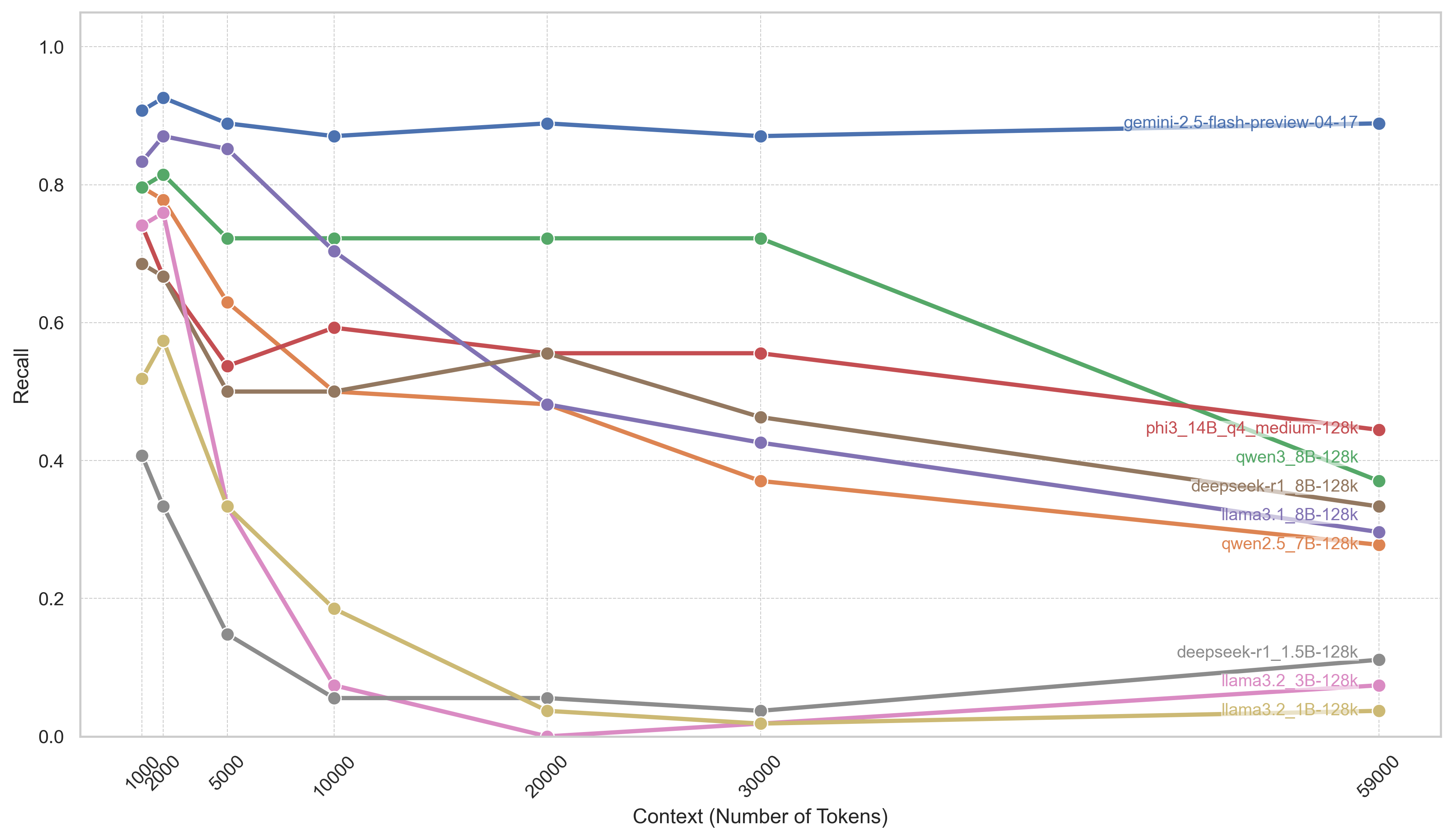}
    \caption{Long-Context QA recall vs. noise: German.}
    \label{fig:zs_recall_noise_de}
\end{figure}

\begin{figure}[H]
    \centering
    \includegraphics[width=\columnwidth]{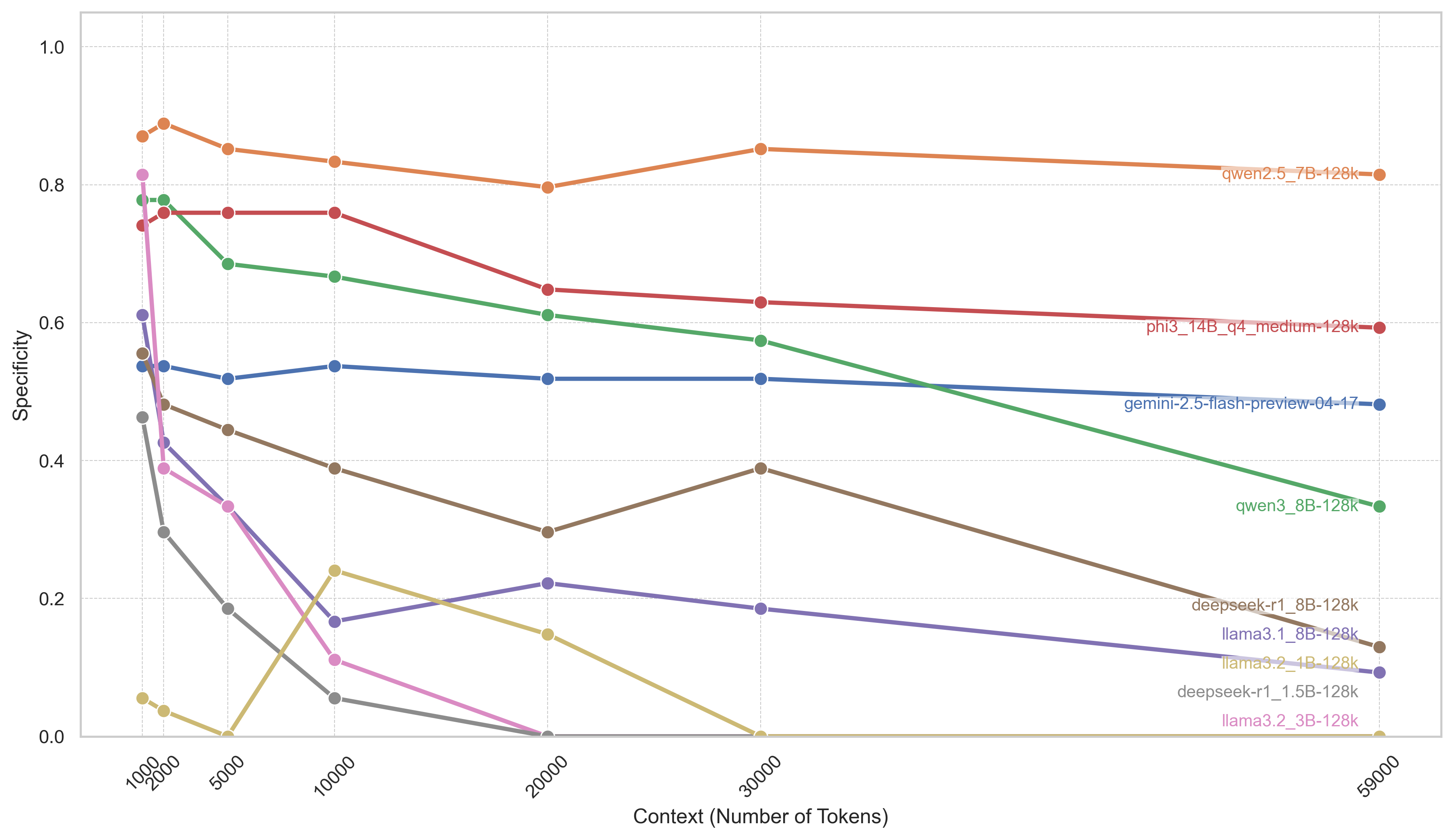}
    \caption{Long-Context QA specificity vs. noise: German.}
    \label{fig:zs_spec_noise_de}
\end{figure}
\clearpage
\onecolumn
\subsection*{Model Performance Comparison}
\begin{figure}[H] 
    \centering
    \includegraphics[width=0.8\textwidth]{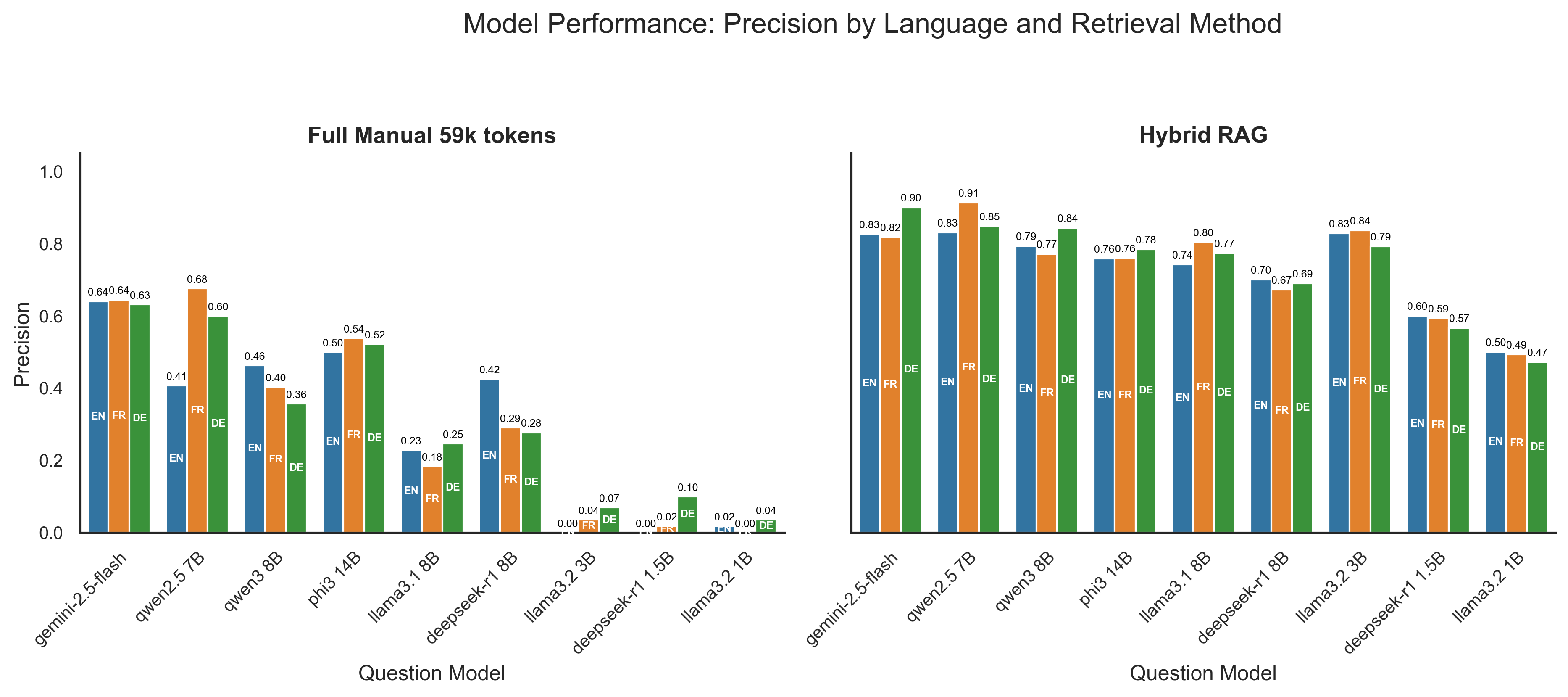}
    \caption{Precision comparison of different models.}
    \label{fig:model_perf_prec_comp}
\end{figure} 

\begin{figure}[H] 
    \centering
    \includegraphics[width=0.8\textwidth]{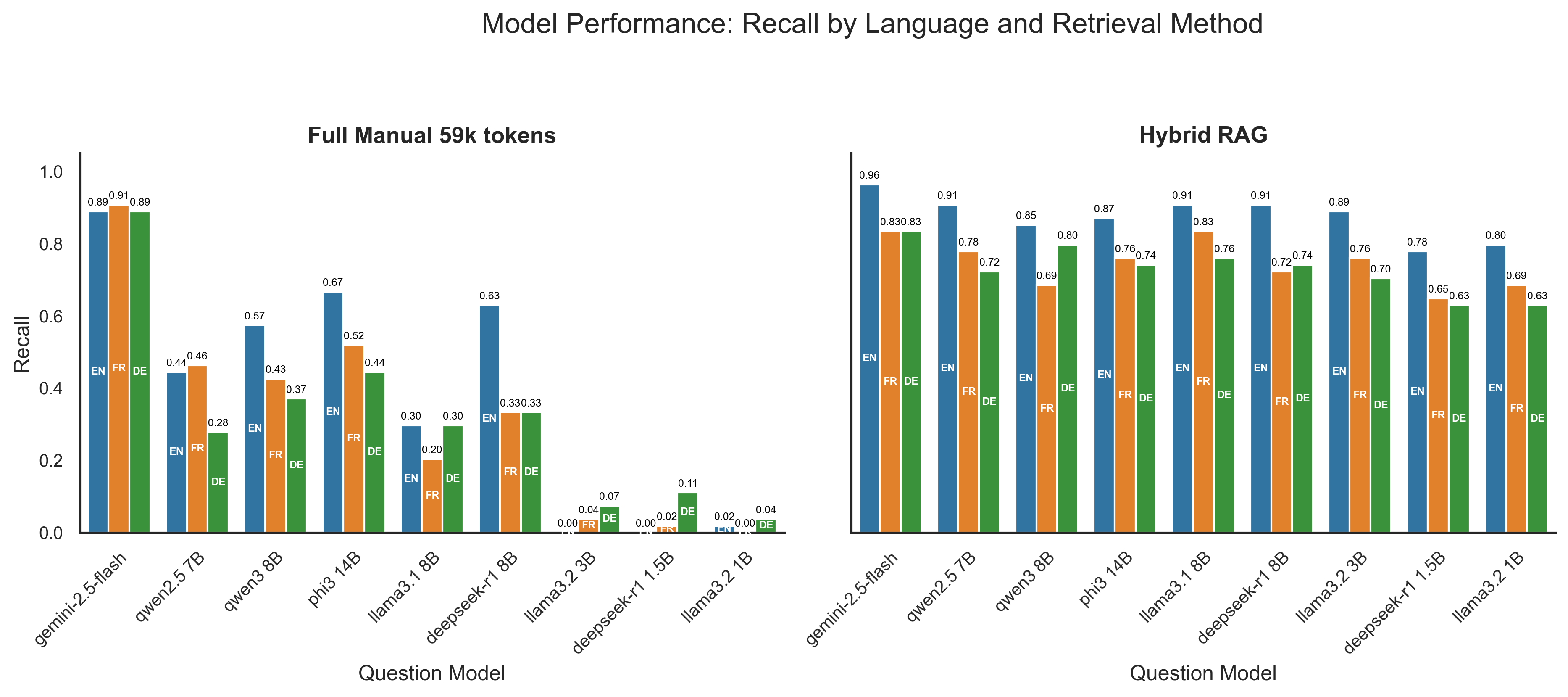}
    \caption{Recall comparison of different models.}
    \label{fig:model_perf_recall_comp}
\end{figure} 

\begin{figure}[H] 
    \centering
    \includegraphics[width=0.8\textwidth]{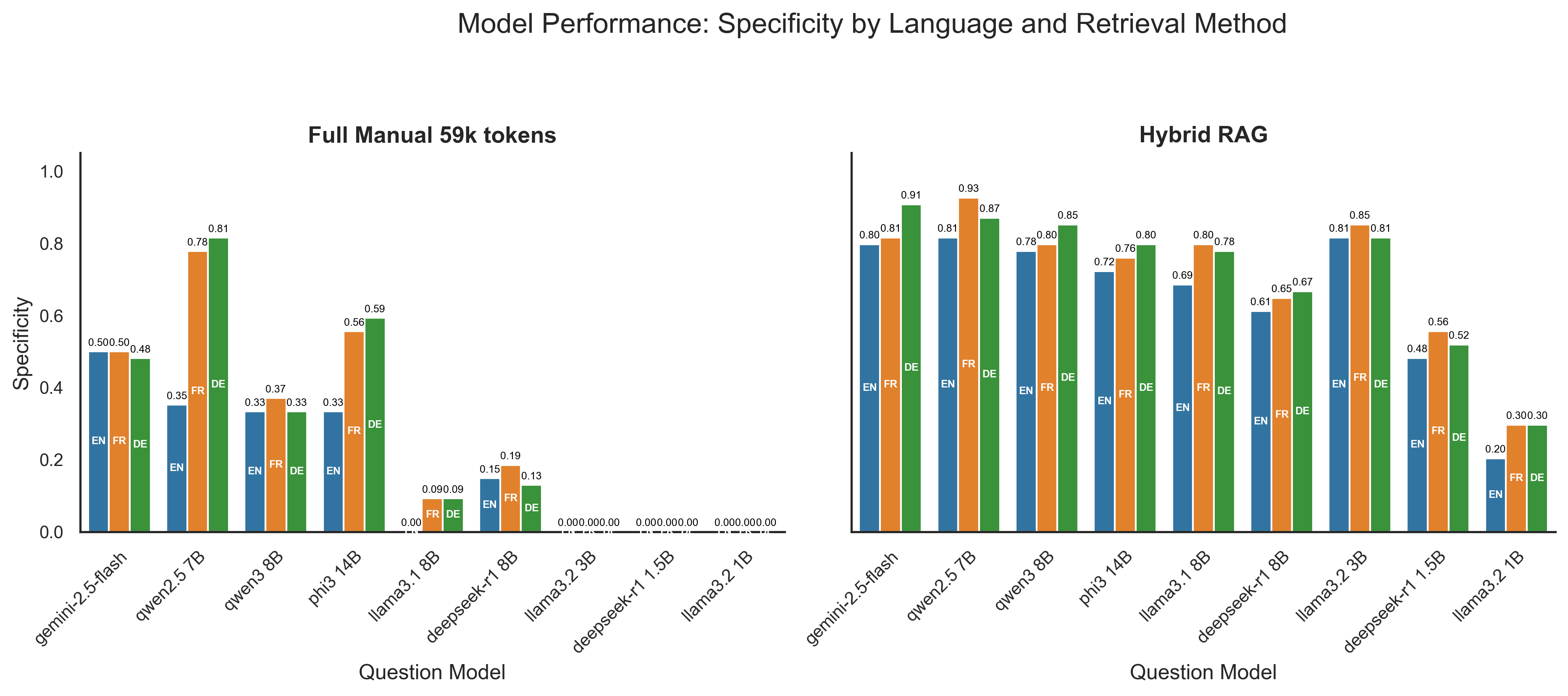}
    \caption{Specificity comparison of different models.}
    \label{fig:model_perf_spec_comp}
\end{figure}

\FloatBarrier 
} 

\clearpage

\onecolumn

\section{Framework Usage Guide}
\label{sec:appendix_framework_usage}

\subsection{Overview}
The benchmarking framework\footnote{The code for our framework is available at \url{https://anonymous.4open.science/r/Agri-Query/}.} is designed to evaluate Large Language Models (LLMs) on technical document understanding. It comprises two main projects: one for Long-Context (LC) testing, often referred to as ``Zeroshot'' testing in the codebase (located in the \texttt{ZeroShot/} directory), and another for Retrieval Augmented Generation (RAG) testing (located in the \texttt{RAG/} directory). Both projects share a common goal: to assess how well LLMs can answer questions based on a provided manual. This appendix provides instructions on data preparation and evaluation execution using both frameworks.

\subsection{Manual Preparation}
The framework primarily ingests manuals in plain text format, often with each page as a separate entry or segment. Manuals in PDF format must be converted to text. The \texttt{ZeroShot} project includes a utility script, \texttt{docling\_page\_wise\_pdf\_converter.py} (located in \texttt{ZeroShot/docling\_page\_wise\_pdf\_converter/}), for this purpose. Executing the main script in the \texttt{ZeroShot} project (\texttt{ZeroShot/main.py}) will automatically attempt to download and convert PDF manuals specified in its configuration, saving them as \texttt{.txt} files. This converter can also be used to prepare text files for the RAG framework.

\subsection{Question Dataset Creation}
Evaluation of LLMs on a new manual requires a corresponding question-answer dataset. This dataset must be a JSON file containing a list of question objects. Each object must include an \texttt{"id"}, \texttt{"question"}, \texttt{"expected\_answer"}, and \texttt{"target\_page"} (the page number in the manual where the answer can be found, or a relevant page for unanswerable questions). For unanswerable questions, the \texttt{"expected\_answer"} should typically be \texttt{"Not found in context"} or a similar designated phrase.

Custom question datasets, for example \texttt{my\_manual\_questions.json}, are placed inside the \texttt{ZeroShot/question\_datasets/} folder for the Long-Context framework. For the RAG framework, the dataset is placed in the \texttt{RAG/question\_datasets/} folder. An example structure for a question entry is shown below:
\begin{appendixpromptstyle}
\begin{verbatim}
// ZeroShot/question_datasets/my_manual_questions.json
// or RAG/question_datasets/my_manual_questions.json

[
  {
    "question": "How many dosing openings are closed during fine application?",
    "answer": "Two of three dosing openings are closed.",
    "page": 24
  },
  {
    "question": "Some question?",
    "answer": "Some answer",
    "page": 99
  }
]
\end{verbatim}
\end{appendixpromptstyle}

\subsection{Long-Context (Zeroshot) Testing Framework}
The Long-Context testing framework, found in the \texttt{ZeroShot/} directory, evaluates an LLM's ability to answer questions when provided with the entire document or large sections of it. This method is also referred to as Zeroshot testing within the project because it tests the model's direct inference capabilities without retrieval augmentation specific to the query. Usage of this framework requires configuration of the \texttt{ZeroShot/config.json} file. This file is used to specify the LLM models, paths to the question datasets, the path or URL to the manual, and other parameters such as noise levels.

The \texttt{main.py} script in the \texttt{ZeroShot/} directory is the entry point for running tests. It is executed from the command line, specifying arguments such as the model, context type, and noise levels. Detailed instructions and configuration options are available in the \texttt{ZeroShot/README.md} file.
An example of relevant parts to update in \texttt{ZeroShot/config.json} for a new manual and dataset:
\begin{appendixpromptstyle}
\begin{verbatim}
// ZeroShot/config.json
{
  "llm_models": { /* ... define models ... */ },
  "evaluator_model": "gemma2:latest",
  "prompt_paths": { /* ... */ },
  "question_dataset_paths": [
    "question_datasets/my_manual_questions.json", // Add new dataset here
    /* ... other existing datasets */
  ],
  // Update document_path or ensure documents_to_test in main.py includes the manual:
  "document_path": "https://yourdomain.com/path/to/your/manual.pdf", // Example for auto download
  /* ... other configurations */
}
\end{verbatim}
\end{appendixpromptstyle}
An example command to run \texttt{ZeroShot/main.py} from within the \texttt{ZeroShot/} directory:
\begin{appendixpromptstyle}
\begin{verbatim}
# From the ZeroShot directory
python main.py --models your_chosen_model --mode all --noise_levels 1000 5000 59000
\end{verbatim}
\end{appendixpromptstyle}

\subsection{Retrieval Augmented Generation (RAG) Testing Framework}
The RAG testing framework, located in the \texttt{RAG/} directory, evaluates LLMs by first retrieving relevant document chunks using various strategies (keyword, semantic, hybrid) and then providing these chunks along with the question to the LLM. Configuration for the RAG framework, including LLM models, embedding models, and dataset paths, is primarily managed through its configuration files (e.g., \texttt{config.ini} or JSON configurations) and command-line arguments for its main evaluation scripts. Manuals must be prepared (e.g., converted to TXT using the \texttt{docling\_page\_wise\_pdf\_converter.py} script from the \texttt{ZeroShot} project and placed in a directory such as \texttt{RAG/manuals/}). The corresponding question dataset must be placed in the \texttt{RAG/question\_datasets/} folder.

The RAG pipeline can be tested with a single question using the \texttt{ask\_question\_demo.ipynb} script, which is typically found within the \texttt{RAG/} directory. This script facilitates inputting a question and specifying the document to observe the retrieved context and the LLM's answer, which is helpful for debugging and exploration before running full-scale evaluations.

For comprehensive evaluations using various RAG strategies and LLMs, the \texttt{RAG/README.md} file provides detailed setup, data preparation (including document processing and vector store creation), and execution instructions for its main evaluation scripts.

\clearpage

\end{document}